%% file: main.tex
\documentclass[10pt,twocolumn,letterpaper]{article}

\input{packages}

\begin{document} 

\input{sections/title}

\input{figures/fig_1_a_banner}

\input{sections/abstract}

\section{Introduction} \label{sec:intro}
\input{sections/introduction}

\section{Related Work}
\label{sec:related}
\input{sections/related_work}

\section{Image Formation in Single-Photon SL}
\label{sec:noisemodels}
\input{sections/noise_model}

\section{Coding for Single-Photon SL}
\label{sec:coding}
\input{sections/coding_schemes}

\subsection{High-throughput Decoding Algorithms}
\label{sec:decoding-al}
\input{sections/decoding}

\section{Experimental Results}
\label{sec:experiments}
\input{sections/experiments}

\section{Conclusion and Discussion}
\label{sec:discussion}
\input{sections/conclusion}

{\small

\input{main.bbl}
}

\onecolumn 
\setcounter{section}{0}
\setcounter{equation}{0}
\setcounter{figure}{0}
\setcounter{table}{0}
\setcounter{page}{1}
\makeatletter
\renewcommand{\theequation}{S\arabic{equation}}

\clearpage
\vspace{-0.1in}
\begin{center}
\textbf{\large Supplementary Material for Single-Photon Structured Light}
\end{center}
\input{supplementary_content.tex}

\end{document}

%% file: packages.tex
\usepackage[pagenumbers]{cvpr}              

\usepackage{graphicx}
\usepackage{amsmath}
\usepackage{amssymb}
\usepackage{booktabs}
\usepackage{comment}

\usepackage{tabularx}
\newcolumntype{L}{>{\raggedright\arraybackslash}X}

\usepackage{caption}
\usepackage{color}
\usepackage{mathtools}
\usepackage{multirow}
\usepackage{multicol}
\usepackage{enumitem}
\usepackage{algorithm}
\usepackage{algpseudocode}

\makeatletter
\usepackage[pagebackref,breaklinks,colorlinks]{hyperref}

\usepackage[capitalize]{cleveref}
\crefname{section}{Sec.}{Secs.}
\Crefname{section}{Section}{Sections}
\Crefname{table}{Table}{Tables}
\crefname{table}{Tab.}{Tabs.}

\def\cvprPaperID{5539} 
\def\confName{CVPR}
\def\confYear{2022}

\makeatother

\newcommand*{\crefex}[1]{%
Suppl. \cref{#1}}

\newcommand{\Prob}[1]{\text{Pr}\left\{#1 \right\}}
\newcommand{\cond}[2]{#1 \,\vert\, #2}

\newcommand{\Set}[2]{%
  \{\, #1 \mid #2 \, \}%
}

\newcommand{\BCH}[3]{\text{BCH}(#1, #2, #3)}
\newcommand{\Repetition}[3]{\text{Repetition}(#1, #2, #3)}

\DeclarePairedDelimiter\ceil{\lceil}{\rceil}
\DeclarePairedDelimiter\floor{\lfloor}{\rfloor}

\newcommand{\bfx}{{\bf x}}

\newcommand{\tildeNice}{{\raise.17ex\hbox{$\scriptstyle\sim$}}}

\newcommand{\approxNice}{{\raise.15ex\hbox{$\scriptstyle\approx$}}}

\newcommand{\ggNice}{{\raise.15ex\hbox{$\scriptstyle >> $}}}

\DeclareCaptionLabelFormat{andfigure}{#1~#2  \&  \figurename~\thefigure}

\DeclareCaptionLabelFormat{andtable}{#1~#2  \&  \tablename~\thetable}

%% file: sections/title.tex
\title{\vspace{-0.2in} Single-Photon Structured Light}

\author{
  Varun Sundar$\,^{\dagger}$ \quad
  Sizhuo Ma$\,^\dagger$ \quad
  Aswin C.\ Sankaranarayanan$\,^\ddag$ \quad
  Mohit Gupta$\,^\dagger$ \\
  {$^\dagger$ University of Wisconsin-Madison} \quad
  {$^\ddag$ Carnegie Mellon University}
}

\maketitle
\renewcommand*{\thefootnote}{$\ast$}
\setcounter{footnote}{1}
\footnotetext{This research was supported in part by the National Science Foundation under the grants 1943149 and 1652569, and the UW-Madison CS summer research award. We also thank Edoardo Charbon for access to the SwissSPAD2 array used in this work.
}

\renewcommand*{\thefootnote}{\arabic{footnote}}
\setcounter{footnote}{0}

%% file: figures/fig_1_a_banner.tex
\begin{figure*}[htp]
    \centering
    \includegraphics[width=\textwidth]{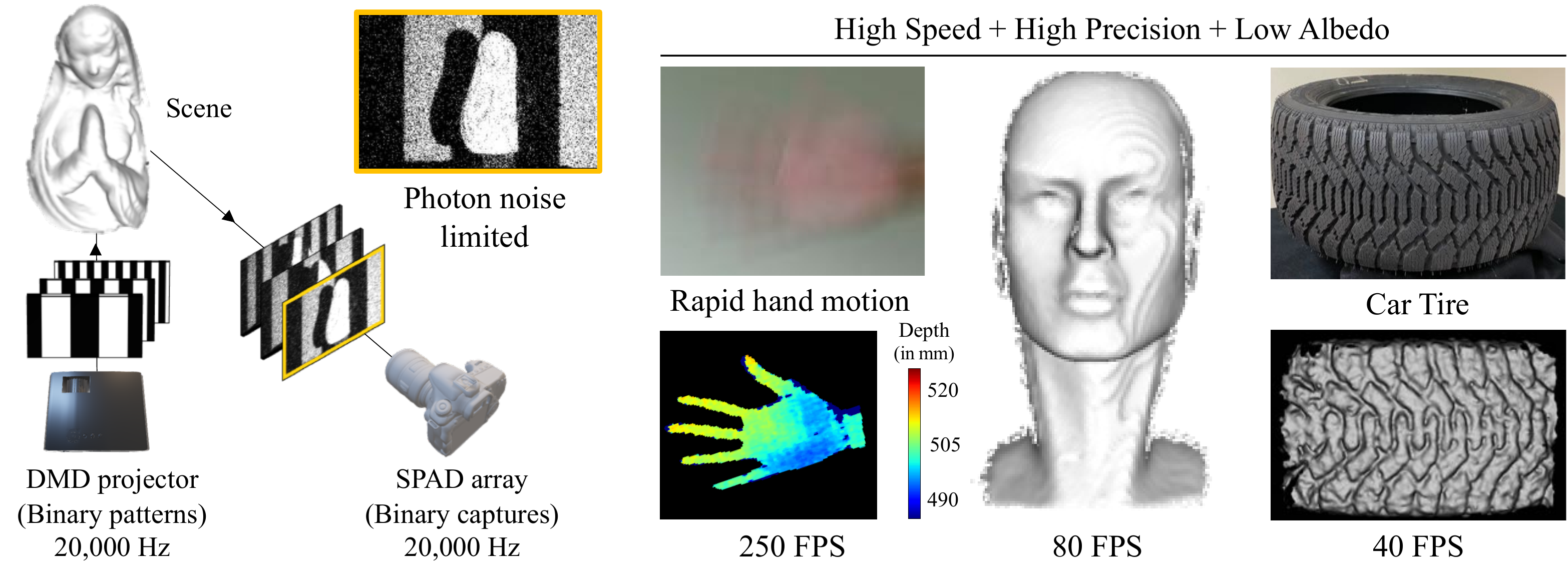}
    \vspace{-0.3in}
    \caption{\textbf{Single-Photon Structured Light.} \textit{(left)} Our proposed system comprises of a SPAD array and a DMD-based projector used to project and acquire binary patterns at extremely high frame rates. We devise coding schemes that can obtain depth-maps from these photon noise limited captures. \textit{(right)} We demonstrate several practical capabilities of Single-Photon SL, including: high-speed scanning of a rapidly moving hand at $250$ FPS, sub-millimeter precise depth-imaging at a range  $50$ cm range and at $80$ FPS, and reconstructing the tread pattern of a tire, a low-albedo object, at $40$ FPS. We include RGB frames captured at $30$ FPS to depict the high speeds involved.}
    \label{fig:banner-intro}
    \vspace{-0.15in}
\end{figure*}

%% file: sections/abstract.tex
\begin{abstract}

We present a novel structured light technique that uses Single Photon Avalanche Diode (SPAD) arrays to enable 3D scanning at high-frame rates and low-light levels.
This technique, called ``Single-Photon Structured Light'', works by sensing binary images that indicates the presence or absence of photon arrivals during each exposure; the SPAD array is used in conjunction with a high-speed binary projector, with both devices operated at speeds as high as 20~kHz.
The binary images that we acquire are heavily influenced by photon noise and are easily corrupted by ambient sources of light.
To address this, we develop novel temporal sequences using error correction codes that are designed to be robust to short-range effects like projector and camera defocus as well as resolution mismatch between the two devices.
Our lab prototype is capable of 3D imaging in challenging scenarios involving  objects with extremely low albedo or undergoing fast motion, as well as scenes under strong ambient illumination.

\end{abstract}

%% file: sections/introduction.tex
Structured light (SL) 3D imaging systems have inherent tradeoffs that balance the precision of the 3D scan against its acquisition time. For instance, temporally-multiplexed SL techniques \cite{guhring2000dense,ha2015multi,GuoMicron} achieve high depth resolution by projecting multiple patterns, thereby precluding high-speed capture. At the other extreme, SL based on spatially-modulated patterns \cite{zhang2010superfast,Hyun:18} can facilitate single-shot scans, but require assumptions of spatial-smoothness that invariably result in loss of detail.

The tradeoffs inherent to SL systems are exacerbated when operating in challenging regimes with low signal-to-noise ratios (SNR) arising from either low-albedo objects, dynamic scenes, or strong ambient illumination. In these scenarios, using longer temporal codes can offer robustness and precision, but at the cost of lowered acquisition speed. One way to mitigate the loss in time resolution is to use high-speed cameras and projectors. However, this approach is limited by large bandwidth requirements and, more fundamentally, the presence of read noise. Each image has a constant amount of read noise, immaterial of the exposure time;  this  can dominate the received signal as the exposure times and, consequently, the image intensities are reduced.

This paper envisions a class of \emph{Single-Photon Structured Light} systems that are based on single-photon detectors, such as Single Photon Avalanche Diodes (SPADs). SPADs can be operated at very high speeds when detecting photons and not their time-of-arrivals. In this `photon detection' mode, the measurements are binary-valued---indicating whether or not a photon arrival occurred during a given acquisition time. For instance, a recently developed SPAD array \cite{ulku512512SPAD2019} can capture \tildeNice $10^5$ binary frames at $1/8$-th megapixel resolution. Our key observation is that the binary measurements, normally considered a limitation due to limited information, are sufficient for a large family of SL coding schemes \cite{posdamer1982surface} that are binary as well.
Since SPADs count photon arrivals, they are not corrupted by read noise. Finally, the use of SPADs for SL finds a natural coupling in high-speed projectors that use digital micromirror devices (DMDs) for displaying binary patterns. \smallskip

\noindent {\bf Coding and decoding for Single-Photon SL.} Due to the probabilistic nature of photon arrivals, the binary-valued measurements captured by SPADs are prone to strong photon noise. For instance, in the presence of strong ambient light, the SPAD could detect a photon even when the corresponding projector pixel is dark. Traditional SL coding schemes are designed for regimes where the image measurements are not binary-valued, and hence are not suitable for Single-Photon SL. We formulate novel SL encoding strategies using error-correction codes that enable robust decoding for Single-Photon SL even under large photon noise. 

Beyond achieving robustness to photon noise, SL coding schemes must account for various imaging phenomena such as projector and camera defocus. Naïve error-correcting codes do not consider these practical effects, and thus cannot be used in a real-world SL system. We design a new class of hierarchical codes using error correction and binary phase shifting that guarantee a minimum stripe width, which provides robustness to such non-idealities. Finally, we design a high-throughput decoding scheme for the proposed codes to enable real-time decoding of the measurements. Our implementation can decode a disparity map for a $512 \times 256$ array in 100 ms on a CPU and 3 ms on a GPU.\smallskip

\noindent {\bf Implications.} Single-photon SL has the potential to enable \emph{extreme 3D imaging capabilities}, including high-speed 3D scanning and robust 3D imaging in low-SNR conditions while respecting low-power and latency budgets. Figure \ref{fig:banner-intro} demonstrates several unique practical capabilities of our prototype Single-Photon SL system, including scanning scenes with low albedo (a tyre) and at high frame rate (fast hand movements) with little loss in the spatial resolution.\smallskip

\noindent {\bf Limitations.} Single-Photon SL inherits limitations endemic to many SL systems. While we mitigate short-range effects such as defocus, long-range effects such as inter-reflections remain to be addressed, possibly by incorporating existing work addressing global illumination. Current SPAD technology is still nascent compared to its CMOS counterparts; the low-resolution of SPAD arrays and their poor fill factors constrains the reconstruction quality of our approach. Fortunately, the capabilities of single-photon sensors continue to improve with higher resolution arrays featuring increased fill factors \cite{morimoto_3.2_2021,Morimoto:20} on the horizon.

%% file: sections/related_work.tex
\noindent {\bf Structured light 3D imaging.} Active triangulation techniques have a rich history with early techniques including stripe scanning \cite{shirai1971recognition,agin1976computer,levoy2000digital}, shadow scanning \cite{bouguet19983d,bouguet19993d}, binary patterns \cite{posdamer1982surface} and sinusoid patterns based phase-shifting~\cite{Srinivasan:85}.  Many methods achieve fast single-shot acquisition by projecting statistical patterns \cite{zhang2012microsoft,Fanello_2016_CVPR,keselman2017intel} or via Fourier Transform Profilometry (FTP)  \cite{An:16,Kemao:04,KEMAO2007304}. Such techniques  require spatial-smoothness assumptions and have low accuracy for strongly textured surfaces.\smallskip

\noindent {\bf Fast binary projectors in SL.} Several SL systems  achieve high-speed 3D scans \cite{zhang2010superfast,Lei:09,zhengPhaseErrorAnalysis2016,su1992automated, Hyun:18,koppal2012exploiting} using the projection capabilities of DMDs.
However, all of these techniques use sensors based on traditional photodiodes which, unlike SPADs, are fundamentally limited by read noise.\smallskip

\noindent {\bf Event-based 3D imaging.} Event-based cameras are bio-inspired devices \cite{gallego_survey_2019} that are triggered asynchronously by intensity changes (or ``events'') typically from a pulsed laser \cite{barndli10.3389/fnins.2013.00275,matsudaMC3DMotionContrast2015,martel2018active} or by a DMD projecting multiple patterns \cite{leroux2018event,mangalore2020neuromorphic,Huang:21}. In contrast to SPADs, event-cameras have 1-2 orders of magnitude lower event density (\tildeNice $10^6$ events/s \cite{Lichtsteiner:2008}). Since each event recovers at most a single 3D point, the limited event density lowers the density and quality of the reconstruction especially in presence of scene-wide motion. While event-based cameras can achieve high dynamic range, their low-light sensitivity remains poor, precluding reliable 3D imaging in low albedo and low SNR scenarios.\smallskip

\noindent {\bf Single-Photon imaging.} Only recently have the capabilities of SPADs, operating without any temporal synchronization, been explored, with applications in high-dynamic range imaging \cite{ingleHighFluxPassive2019,inglePassiveInterPhotonImaging2021} and burst photography \cite{maQuantaBurstPhotography2020}. Our method operates the SPAD array similar to Ma et al. \cite{maQuantaBurstPhotography2020}, using a sequence of binary frames. Although we focus on SPADs due to their superior frame-rate, the proposed techniques are applicable to other single-photon imaging technologies such as Jots \cite{fossum2005sub,fossum2016quanta}, which feature high-resolution arrays with smaller pixels and increased photon-efficiency \cite{Ma:17}, albeit at lower frame-rates and higher read noise. 

%% file: sections/noise_model.tex
Consider a SPAD pixel array observing a scene. The number of photons $N$ arriving at a pixel $\bfx$ during an exposure time  $t_\text{exp}$ is modelled as a Poisson random variable:
\begin{equation}
\label{eq:Poisson_model}
\begin{aligned}
    \Prob{N = k} = \frac{(\Phi(\bfx) \, t_\text{exp})^k \,\, e^{-\Phi(\bfx) \, t_\text{exp} }}{k!} \,,
\end{aligned}
\end{equation}
where $\Phi(\bfx)$ is the flux\footnote{For simplicity, we assume a $100\%$ quantum efficiency and use the term ``flux'' interchangeably with the  arrival rate of photo-electrons.}. During each exposure, a pixel detects at most one photon, returning a binary value $B(\bfx)$ such that $B(\bfx) = 1$ if the pixel detects one or more photons. Hence, $B(\bfx)$ is a Bernoulli random variable \cite{yang_poisson_model} with
\begin{equation}
\label{eq:shot_noise_model}
\begin{aligned}
    \Prob{B(\bfx) = 0} = e^{-\left({ \Phi(\bfx)   + r_q}\right)t_\text{exp}},  
\end{aligned}
\end{equation}
where $r_q$ is the dark current rate---the rate of spurious counts unrelated to incident photons.

In a typical SL scan, the scene is illuminated with a sequence of 2D binary patterns from a projector.
The SPAD captures a binary frame for each pattern. Each SPAD pixel receives a binary code over time, from which we estimate the projector column observed at the pixel---an operation that is critical for the success of any SL technique.

We now derive the probability that a projected temporal sequence will be decoded incorrectly. 
For a given binary pattern, consider a SPAD pixel $\bfx$ that observes a scene point illuminated by an ON pixel. Suppose the incident photo-electron arrival rate at $\bfx$ due to projector and ambient illumination are $\Phi_p(\bfx)$ and $\Phi_a$, respectively. Then, the probability of a bit-flip error, i.e., the probability of the SPAD pixel not detecting a photon is given as 
\begin{equation}
\text{P}_\text{flip, bright} = \Prob{\cond{B(\bfx)=0}{\Phi(\bfx)=\Phi_a+\Phi_p(\bfx)}} \,.
\label{eq:flipbright}
\end{equation}
Similarly, the probability of a bit-flip error when the projector pixel is OFF, i.e., the probability of detecting a photon in spite of not illuminating the corresponding projector pixel is 
\begin{equation}
\text{P}_\text{flip, dark} = \Prob{\cond{B(\bfx)=1}{\Phi(\bfx)=\Phi_a}}.
\label{eq:flipdark}
\end{equation}
The bit-flip probabilities for ``bright'' and ``dark'' pixels, 
visualized in \cref{fig:bit-flip} for varying $\Phi_a$ and $\Phi_p$, are not equal due to the asymmetric role played by the ambient photons. 
\input{figures/fig_2_bit_flip}

We can now compute the probability of incorrectly decoding of an $L$-length Gray code. Since it is equally likely to observe any $L$-bit binary code, the average probability of erroneous decoding over all codewords is:
\begin{equation}
\label{eq:gray_code_error}
\Prob{\text{error}} = 1 - \left( 1 - \left( \frac{\text{P}_\text{flip, bright}+\text{P}_\text{flip, dark}}{2}\right) \right)^{L}.
\end{equation}
A detailed derivation is provided in \crefex{supp_sec:gray_deriv}.\smallskip

\noindent{\bf Typical decoding error probabilities.} \Cref{fig:prob} shows the decoding error probability for a $10$-bit Gray code across ambient light levels. Increasing ambient light levels drastically increases $\text{P}_\text{flip, dark}$, resulting in a near-certain decoding failure. In the next section, we propose coding strategies for Single-Photon SL that enable accurate decoding even in highly challenging conditions.

\input{figures/fig_2_b_table_fig}

%% file: figures/fig_2_bit_flip.tex
\begin{figure}[ttt]
    \centering
    \includegraphics[width=\columnwidth]{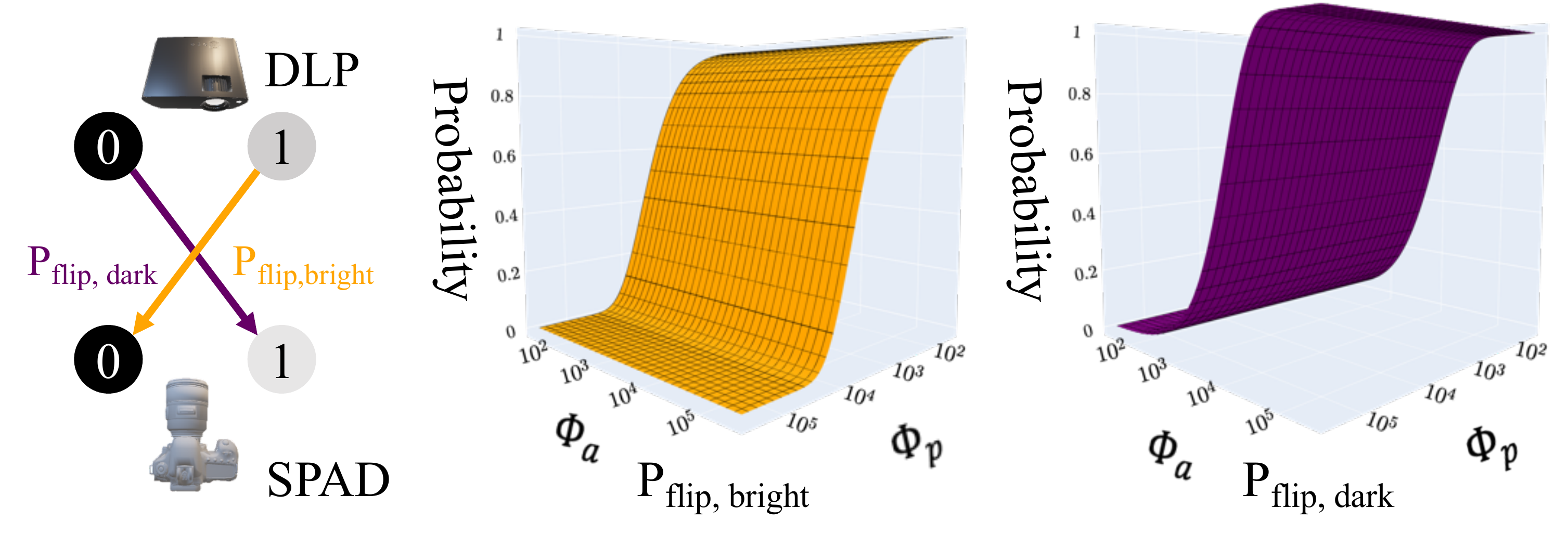}
    \vspace{-0.3in}
    \caption{\textbf{Single-Photon SL features an asymmetric noise model.} Bit-flip probabilities as determined by Eqs. (\ref{eq:flipbright}) and (\ref{eq:flipdark}) are evaluated across a grid of $(\Phi_a, \Phi_p)$ flux values. The plot parameters are $t_\text{exp}=10^{-4}s$ and dark current rate $r_q=0$.}
    \label{fig:bit-flip}
    \vspace{-0.15in}
\end{figure}

%% file: figures/fig_2_b_table_fig.tex
\begin{figure}
  \centering
  \begin{subfigure}{\columnwidth}
    \centering
    \input{tables/bit_flips}
    \caption{Bit-flip probabilties}
  \end{subfigure} \\
  \vspace{0.05in}
  \begin{subfigure}{\columnwidth}
    \includegraphics[width=\columnwidth]{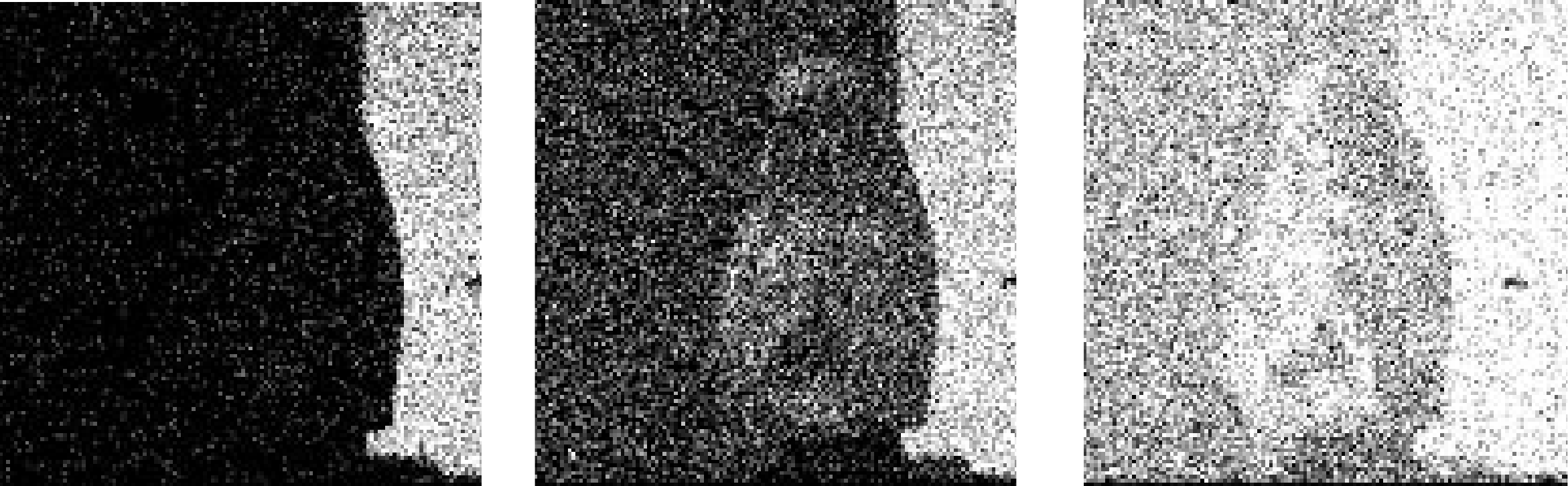}
    \caption{Binary frames captured by Single-Photon Camera}
  \end{subfigure}
  \vspace{-0.1in}
  \caption{\textbf{Typical bit-flip probabilities} observed in our lab prototype under different ambient illumination conditions at an exposure of $10^{-4}$s. (a) Using these probabilities, we can compute the decoding error probability for a code sequence of length $L=10$. (b) We project a pattern where the left half is dark and the right half is bright. At higher ambient intensities, more photons are detected in the ``dark'' region---which makes reliable decoding a challenge.}
  \label{fig:prob}
  \vspace{-0.15in}
\end{figure}

%% file: tables/bit_flips.tex
\begin{tabular}{@{}cccc@{}}\toprule
& $\text{P}_\text{flip, dark}$ & $\text{P}_\text{flip, bright}$ & $\Prob{\text{error}}$ \\
\midrule
dark room & 0.021 & 0.22 & 0.72\\
indoor lamp & 0.23& 0.19 & 0.9\\
spot lamp & 0.75&  0.06&  0.99\\
\bottomrule
\end{tabular}

%% file: sections/coding_schemes.tex
\input{figures/fig_method_overview}

We now describe temporal coding schemes for Single-Photon SL with the goal of achieving robustness to random bit-flips using error correction mechanisms, and incorporating practical considerations in code design. The overall coding and decoding pipeline is illustrated in \cref{fig:method-overview}.

One simple strategy to improve the reliability of any scheme is to repeat the projected patterns and perform a majority vote. This seems to be a viable option since the high-speed projection and capture of DMDs and SPADs, respectively, affords high temporal redundancy. For example, given a code sequence of length $L=10$, we could simply repeat the patterns $25$ times (called the redundancy factor), and still maintain a high overall frame rate for 3D capture. Such a majority vote will improve the decoding performance, provided the probability of bit flips is less than $0.5$. However, in extreme conditions (say, low SNR) a large redundancy factor may be needed to achieve even a modest improvement. 

Can we design coding schemes for Single-Photon SL that perform better than simple repetition? 
It is well-known in coding theory that repetition is a sub-optimal error-correcting mechanism \cite{roth2006introduction}. We, instead, turn to a popular family of binary error-correcting codes, the Bose–Chaudhuri–Hocquenghem (BCH) codes \cite{bose1960class}, used in applications ranging from QR codes \cite{QRCodes} to satellite communication \cite{bch_satellite}. Our choice of BCH is motivated by their ability to correct stochastic errors and their flexibility in designing codes with varying redundancy factors---which permits a graceful tradeoff  between speed and robustness. 

\subsection{BCH Codes}
\label{sec:error-correcting}

 A $\BCH{n}{k}{d}: \{0,1\}^k \to \{0, 1\}^n$ encoder takes input messages of length $k$ and produces output codewords of length $n$ that are at least $d$-bits apart.
 Hence, such a coding scheme provides error correction capabilities up to $\floor{\frac{d-1}{2}}$ bit flips, in the worst case or adversarial sense. 
%
This worst-case error-correcting capacity of BCH codes is significantly higher than of the repetition code.
For instance, $\BCH{63}{10}{27}$---which uses 63-length codewords to encode messages of length 10---can correct up to $13$ worst-case errors, while the corresponding capability for $\Repetition{60}{10}{6}$, where the message pattern is repeated $6$ times, can correct only $2$ errors.
Going further, in our problem setting, the main source of bit flips is photon noise, which is stochastic and non-adversarial, and thus we can expect error correction beyond the worst-case regime.\smallskip

\noindent {\bf Designing BCH-encoded patterns.} Consider a projector with $C$ columns. We aim to design projector patterns so that each column is assigned a unique binary code with in-built BCH error correction. To produce the projector patterns, we start with a base binary coding scheme that uniquely represents each projector column, for example, with Gray codes~\cite{Inokuchi:1984}. 
Given a set of message codes $\{ m_i\} \subseteq \{0, 1\}^L$, where $L=\ceil{\log_2 C}$, we choose a BCH encoder $\mathcal{E}_{\BCH{n}{k}{d}}$ that is capable of encoding at least $C$ messages (i.e., $k \geq L$) and output column-wise codes $\{ c_i = \mathcal{E}_{\BCH{n}{k}{d}}\left(m_i \right) \}$.

As an example, \cref{fig:bch-31-11} illustrates the \BCH{31}{11}{11} encoding of 10-bit Gray code messages. Since $L < k$ here, we use \textit{shortening}, i.e., we prepend the message by $(k-L)$ zeros, but do not transmit them, thereby reducing the projected code length from $n$ to $n - (k - L)$. 
%
We also use \textit{systematic encoding}, i.e., the first $L$-bits of each code is the message itself---hence each sequence comprises of message patterns appended by parity patterns.
The choice of $n$, the length of the BCH code, simultaneously determines the robustness of the code as well as the loss in time resolution. Longer codes have better error-correction capabilities; but since we need to acquire a larger temporal sequence, this reduces our ability to handle fast(er) moving objects. With this in mind, in the rest of the paper, we present results at two operating points---$n=\{ 63, 255\}$---to cover two distinct scenarios.\smallskip

\input{figures/fig_4_bch_31_11_5}

\noindent {\bf Evaluating BCH encoding for Single-Photon SL.} To understand the benefits of BCH encoding, we use  Monte-Carlo simulation of decoding error probability across a grid of $(\Phi_a, \Phi_p)$ values, and compare conventional Gray codes, repetition codes and BCH codes for 10-bit binary messages. The performance of these schemes is presented in  \cref{fig:bch-repetition}. At most operating points, the decoding error probability of BCH codes is either close to zero or presents an order of magnitude improvement over repetition. 

\subsection{Code Design under Practical Considerations}
\label{sec:practical-cons}
Beyond achieving robustness to photon noise, SL coding schemes must account for various imaging non-idealities. A majority of the BCH encoded patterns, as seen in \cref{fig:bch-31-11}, comprise of high-spatial frequency patterns that do not perform well under projector/camera defocus and resolution mismatch between the devices.
Therefore, in spite of achieving low errors in theory, BCH codes as described so far will simply be inadequate in a practical SL system.

A common approach to mitigate such short-range effects is to use long-run Gray codes, which are a subset of Gray codes that maximize the shortest stripe width \cite{guptaStructuredLight3D2011,goddyn1988gray,goddynBinaryGrayCodes2003} across all the projected patterns. However, applying BCH encoding on long-run Gray codes also results in a majority of patterns containing high-spatial frequencies (see \crefex{fig:supp-bch-31-long-run}). Finding binary messages $\{m_i\}$ that maximize the minimum stripe-width of BCH patterns $\{c_i\}$ is an intractable combinatorial problem with an exorbitant solution space ($~1024!$ candidate solutions).\smallskip

\noindent{\bf Hybrid codes.} Our key idea is to design hierarchical codes where BCH encoding is performed only on the more significant bits (MSBs) of the base Gray code pattern. This ensures that all the BCH-encoded frames have large minimum stripe widths, making them robust to defocus effects. The remaining lower significant bits (LSBs) are resolved using circularly-shifted binary patterns, where we shift the pattern one-pixel-at-a-time to represent columns sequentially. We term this as ``binary shifting''. 

Specifically, given a $L$-bit message, we encode its $L_\text{BCH}$ MSBs and the remaining $L_\text{shift} = L - L_\text{BCH}$ LSBs in different ways.
The MSBs are coded using BCH as described earlier; since the message codes corresponding to a specific MSB pattern remains unchanged for all values of the LSBs, the resulting BCH codes have a stripe width of at least $2^{L_\text{shift}}$. 
The $L_\text{shift}$ LSBs are coded by a temporal sequence of length $2^{L_\text{shift} + 1}$ featuring a burst of $2^{L_\text{shift}}$ ones---whose starting position (or phase) encodes the message.
\Cref{fig:hybrid-repetition} illustrates the codewords arising for this hybrid construction, which are guaranteed to have an overall stripe width of at least $2^{L_\text{shift}}$ pixels. 
In our implementation, for a $L=10$-bit message, we set $L_\text{BCH}=7$ and $L_\text{shift}=3$, featuring a minimum stripe width of $2^3 = 8$ pixels. Additionally, we utilize BCH encoders with $n \in \{63,255\}$. 

\input{figures/fig_3_bch_repetition_comparison}

We note that this coding scheme is similar in spirit to hybrid SL techniques \cite{chenModifiedGrayLevelCoding2017,gorthi2010fringe,zhengPhaseErrorAnalysis2016,zhengPhaseshiftingProfilometryCombined2017} where Gray codes provide global disambiguation and Phase Shifting \cite{guhring2000dense} resolves LSBs, providing precise correspondences. However, binary shifting has a key difference compared to phase shifting in that intensity information cannot be inferred from a binary measurement. Consequently, unlike phase shifting, where a single measurement can determine the unwrapped phase, binary shifting requires projecting multiple patterns.

Binary shifted patterns are decoded using a matched filter approach, by autocorrelating the received sequence with an unshifted stripe sequence. 
In \crefex{supp_sec:binary-shift}, we show that binary shifted patterns offer significant robustness to random bit-flips by deriving the expected decoding error. 
Finally, to illustrate the hybrid codes' overall error-correcting capability, we compare them to repeated long-run Gray codes of similar codelength. 
We characterize performance using root mean squared error (RMSE) in decoded correspondence as the error metric. 
As we observe in \cref{fig:hybrid-repetition}, hybrid codes outperform repeated Gray codes, and the performance gap increases at higher redundancy factors.

\input{figures/fig_5_hybrid_bch}

%% file: figures/fig_method_overview.tex
\begin{figure*}[htp]
    \centering
    \includegraphics[width=\textwidth]{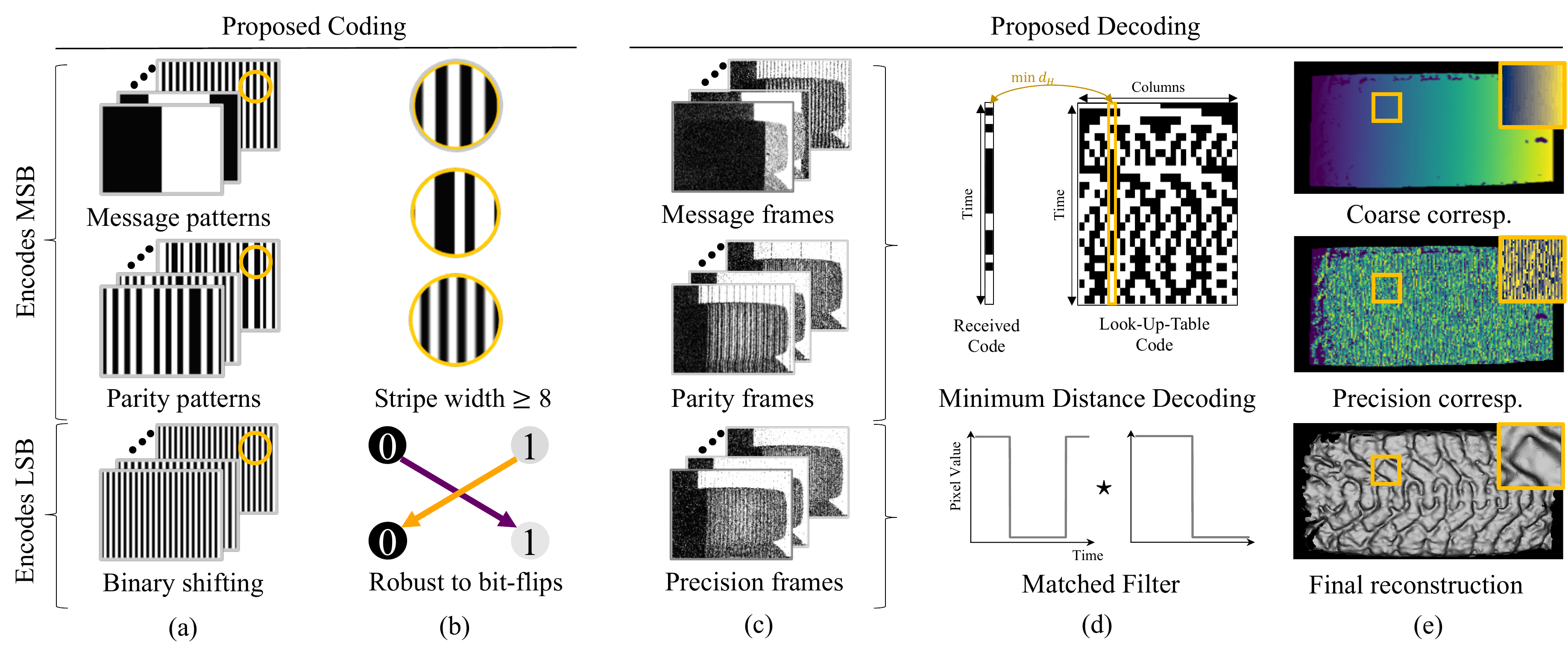}
    \vspace{-0.35in}
    \caption{\textbf{Overview of Single-Photon SL.} (a) Our coding scheme assigns to each projector column a unique binary message. The most significant bits (MSBs) of this message are transformed using error-correcting encoders (\cref{sec:error-correcting}), while the least significant bits (LSBs) are encoded using circularly shifted binary patterns (\cref{sec:practical-cons}). (b) This \emph{hybrid} strategy offers robustness to bit-flips arising from photon noise while guaranteeing a minimum stripe-width, thereby retaining its effectiveness even in the presence of short-range non-idealities. (c) Captured frames, acquired here in $25$ms, feature a large amount of photon noise.  (d) Similar to encoding, the MSBs and LSBs portions are decoded separately, using minimum distance decoding and a matched filter respectively. In \cref{sec:decoding-al}, we present a high-throughput decoding procedure that outputs correspondences in real-time. (e) Correspondence maps and the final 3D reconstruction illustrate the coarse-to-precision approach of the {Hybrid} strategy. Zoomed-in insets of correspondence maps are plotted with a different colormap.}
    \label{fig:method-overview}
    \vspace{-0.15in}
\end{figure*}

%% file: figures/fig_4_bch_31_11_5.tex
\begin{figure}[ttt]
    \centering
    \includegraphics[width=\columnwidth]{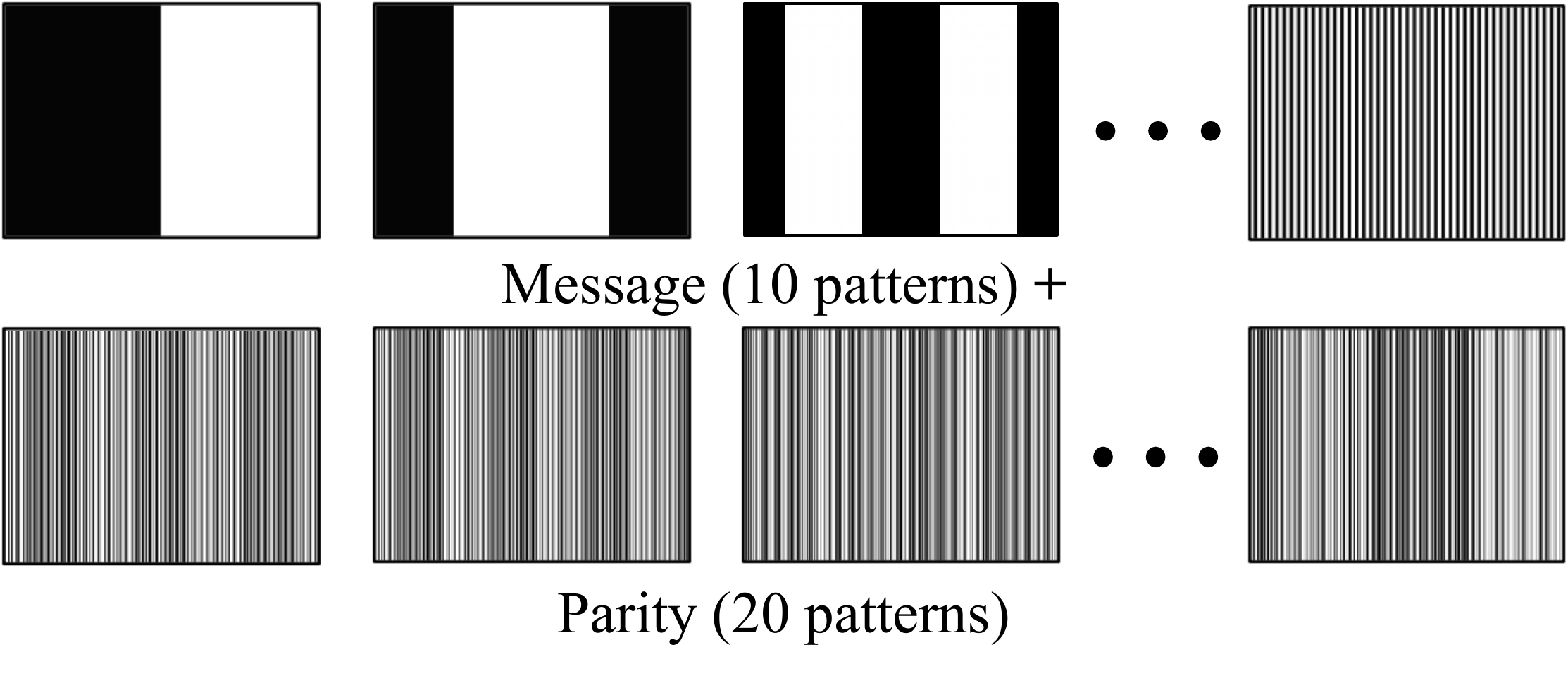}
    \vspace{-0.3in}
    \caption{\textbf{\BCH{31}{11}{11} encoding of a 10-bit Conventional Gray message.} We use \textit{systematic} encoding, where message patterns are appended by parity patterns, providing tolerance to random bit-flips caused by photon noise. The complete code lookup table describing these patterns is shown in \crefex{fig:supp-bch-31-gray}.}
    \vspace{-0.1in}
    \label{fig:bch-31-11}
\end{figure}

%% file: figures/fig_3_bch_repetition_comparison.tex
\begin{figure}[ttt]
    \centering
    \includegraphics[width=\columnwidth]{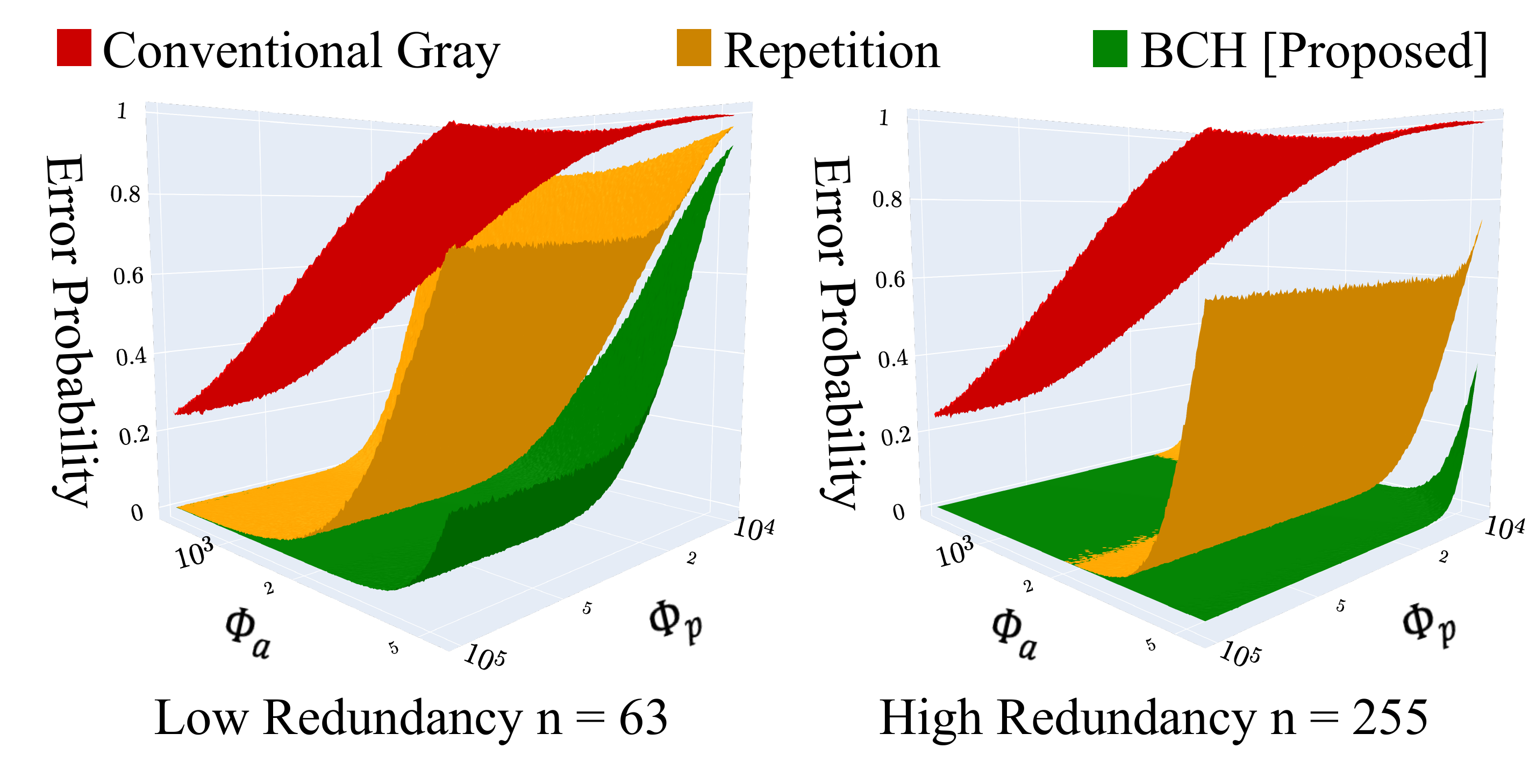}
    \vspace{-0.3in}
    \caption{\textbf{Monte-Carlo evaluation of BCH and repetition strategies.} We empirically evaluate the probability of decoding error upon receiving a codeword randomly corrupted by bit-flips. The ambient flux ($\Phi_a$) and projector flux ($\Phi_p$) values at a pixel location determine the bit-flip probability. We use BCH encoders with $n=\{63, 255\}$. Both repetition and BCH strategies improve the robustness of conventional Gray codes to photon noise. Additionally, BCH outperforms repetition at all $(\Phi_a,\Phi_p)$ with a pronounced difference at higher redundancies.}
    \label{fig:bch-repetition}
    \vspace{-0.1in}
\end{figure}

%% file: figures/fig_5_hybrid_bch.tex
\begin{figure}[ttt]
    \centering
    \includegraphics[width=\columnwidth]{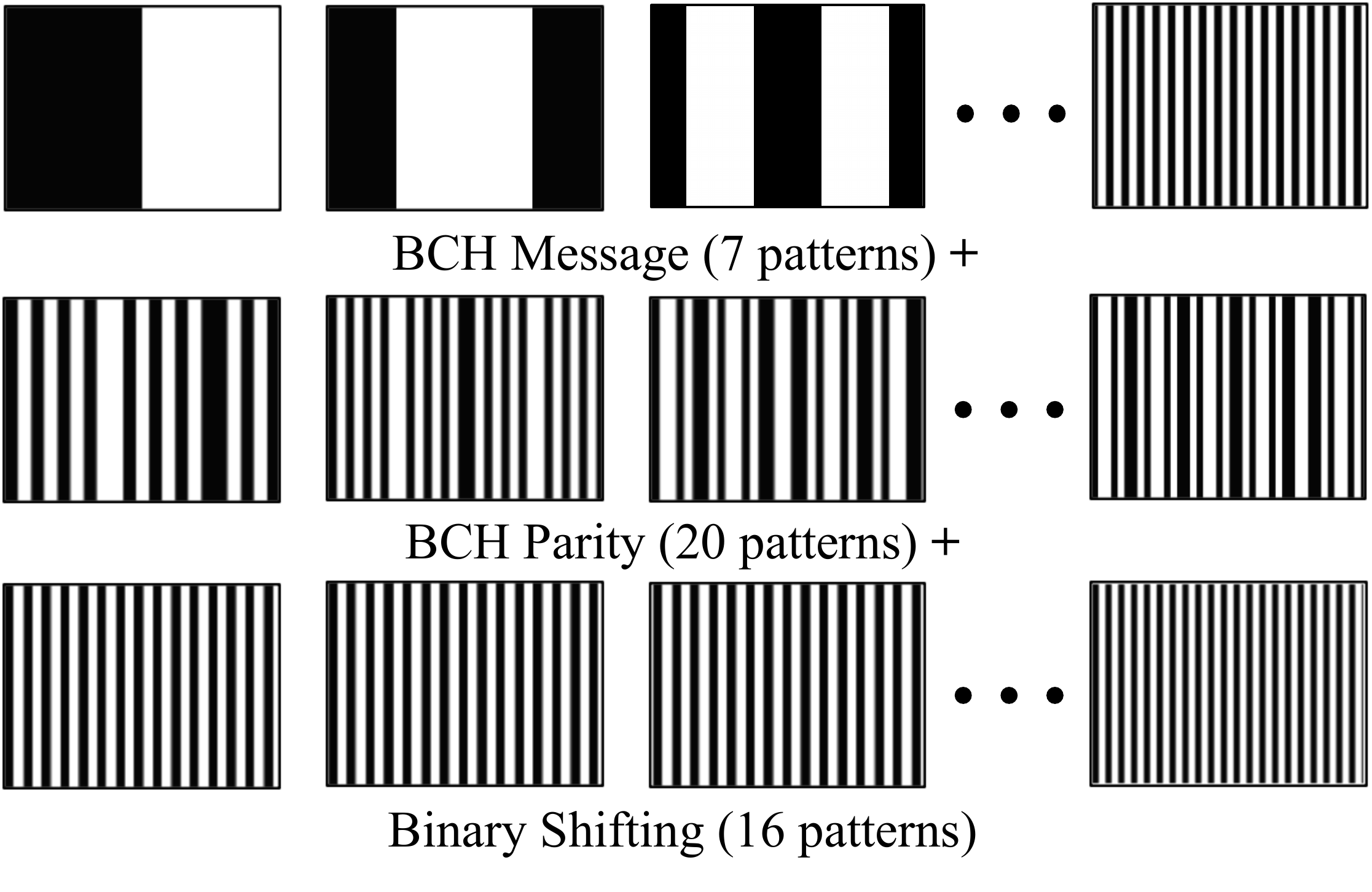}
    \vspace{-0.3in}
    \caption{\textbf{{Hybrid} encoding of a 10-bit conventional Graycode message}. We use BCH encoding for the first $7$ bits of the message and binary shift encoding for the last $3$ bits, resulting in a minimum stripe width of $8$ pixels. Hence, these patterns do not possess high spatial-frequency, and are more robust to short-range effects.} 
    \label{fig:hybrid-bch-31}
    \vspace{-0.1in}
\end{figure}

%% file: sections/decoding.tex
In applications such as robotic navigation, augmented reality and high-throughput industrial inspection, it is imperative to obtain depth maps at speeds comparable to frame acquisition. For conventional codes, like Gray and repetition codes, decoding can be performed via fast analytical algorithms. While analytical decoding methods (e.g., the Berlekamp Massey algorithm \cite{berlekamp_massey}) with polynomial run-time exist for BCH codes as well, these methods can only correct up to the worst-case error (Hamming) limit, which is insufficient due to a potentially large number of bit-flips caused by photon noise.
In this section, we discuss fast decoding techniques for BCH and, as an extension, the hybrid codes proposed in Sec.\ \ref{sec:coding}. The goal is to design techniques that can achieve real-time decoding, while also being able to handle a large number of individual bit-flip errors.\smallskip

\noindent{\bf Minimum distance decoding for Single-Photon SL.} One simple decoding approach is Minimum Distance Decoding (MDD), where the measured codewords are compared against every projected code word. MDD, while conceptually simple, can correct errors beyond the worst-case limit\footnote{The exact number of  correctable errors depends on $(\Phi_a, \Phi_p)$; with MDD being the Maximum Likelihood Decoder when $\text{P}_\text{flip, bright}$ \approxNice  $\text{P}_\text{flip, dark}$.}.
However, a brute-force implementation of MDD can often be unviable, owing to its exorbitant run-time and/or memory requirements.
Fortunately, Single-Photon SL has certain favourable properties that lead to a fast, high-throughput MDD procedure. First, the space of messages (number of projector columns, \tildeNice $2^{10}$) is significantly smaller than space of codewords ($2^n$, $n \in \{63, 255\}$). Second, the number of queries for decoding, which is the number of pixels in the SPAD sensor,  exceeds the number of messages. 

\input{figures/fig_6_hybrid_vs_repeat}

These circumstances permit us to leverage the recent progress in similarity search  \cite{aumuller2017ann,shimomura2021survey}, which has lead to efficient nearest-neighbour algorithms for batched queries. Based on  empirical comparisons (presented in \crefex{fig:supp-bench}), we find that 
FAISS \cite{faiss_johnson2019billion} offers the highest throughput, decoding a $1/8$th MP array in 100 ms on CPU and 3 ms on GPU. Such methods also scale to larger arrays, requiring $12$ ms and $30$ ms for one and four megapixels respectively.

%% file: figures/fig_6_hybrid_vs_repeat.tex
\begin{figure}[ttt]
    \centering
    \includegraphics[width=\columnwidth]{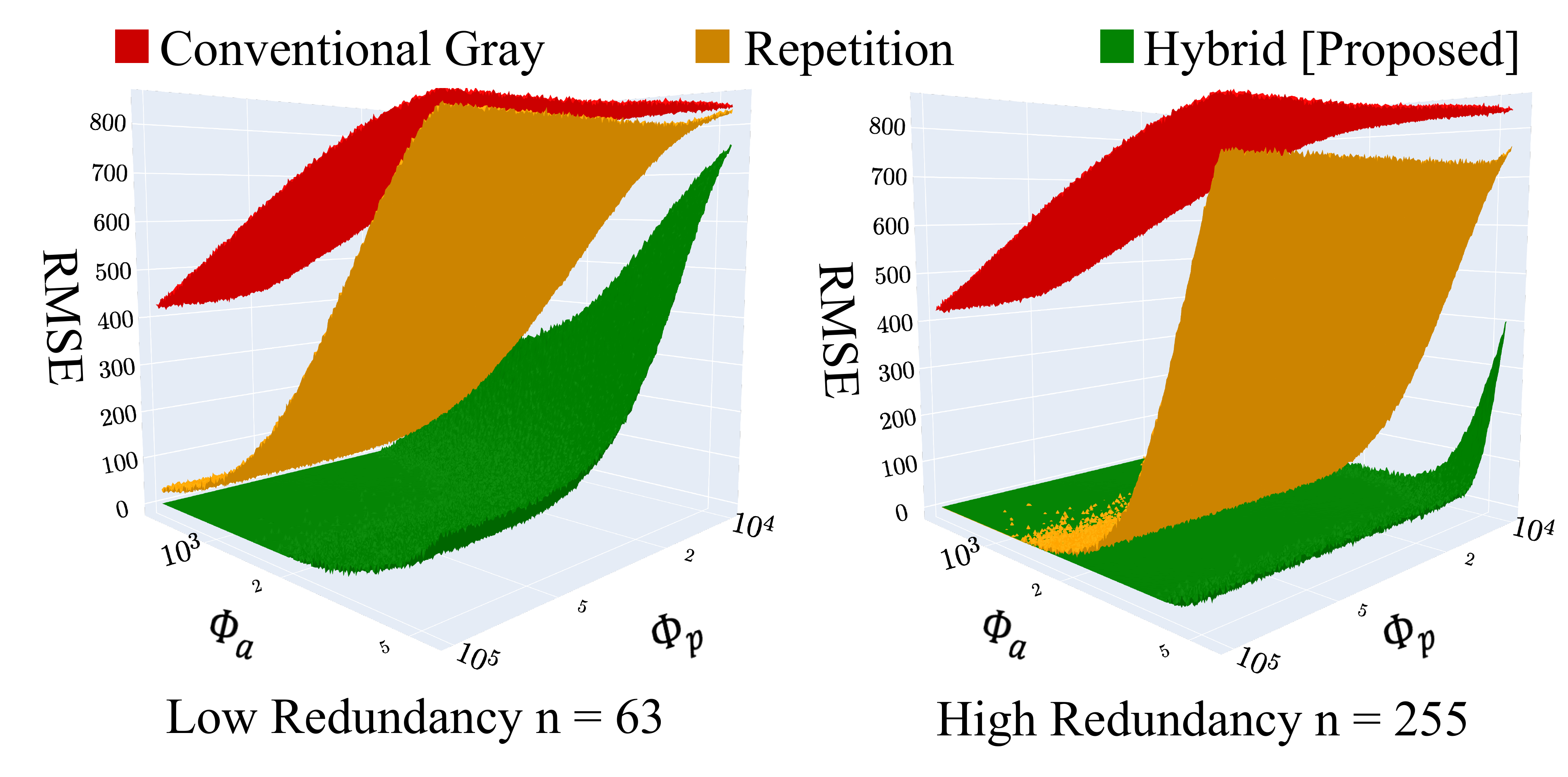}
    \vspace{-2em}
    \caption{\textbf{Hybrid codes are more robust to photon noise than repeated Gray codes, evaluated across a grid of $(\Phi_a, \Phi_p)$ photon fluxes.} We are interested in patterns robust to projector defocus and resolution-mismatch effects and hence consider repetition with long-run Gray codes. For a 10-bit message, both repeated long-run Gray and Hybrid codes have a minimum stripe width of $8$ pixels. We use RMSE (in correspondence) as the evaluation metric to account for the locality in decoding error.}
    \vspace{-0.1in}
    \label{fig:hybrid-repetition}
\end{figure}

%% file: sections/experiments.tex
\input{figures/fig_11_method_comp_unified}

We now describe a range of experiments to demonstrate the performance of Single-Photon SL. Our lab prototype was constructed with the SwissSPAD2 array \cite{ulku512512SPAD2019}, which is a $512 \times 256$ SPAD array. The array has a pixel pitch of $16.38$~$\mu\textrm{m}$, and can capture binary frames at speeds up to $100$ kHz. In \crefex{supp_sec:hardware}, we provide additional details regarding the setup including the calibration procedure used.

\subsection{Single-Photon SL on Static Scenes}
To characterize the performance of Single-Photon SL, we image static scenes of varying albedo and ambient light levels.
We use these case studies to compare the performance of different error correction schemes and to show the effectiveness of the proposed hybrid codes.
To obtain ground truth scans, we operate a DMD projector at a low frame-rate of 2 Hz, while running the SPAD at $10240$ Hz, thereby obtaining $5120$ SPAD frames per projected pattern.
The average of $5120$ frames has minimal photon noise, and is considered as a ground truth measurement.\smallskip

\noindent{\bf Performance of proposed codes.} \Cref{fig:method-comp-un} compares our proposed Hybrid and BCH strategies to repeated Gray codes and repeated long-run Gray codes. We report overall RMSE and RMSE among inliers, thereby measuring both accuracy and consistency. As seen in \cref{sec:practical-cons}, the proposed hybrid codes are considerably more consistent and accurate across the two redundancy factors used.
Whereas, naïve BCH codes are heavily distorted due to defocus. Since each strategy has access only to a single binary frame per projected pattern, this emulates a \emph{3D capture speed} of $40$ FPS in Hybrid $(n = 255)$ and $130$ FPS in Hybrid $(n = 63)$.\smallskip

\input{figures/fig_12_ambient_light}

\noindent{\bf Low SNR regimes.} \Cref{fig:ambient-light} examines reconstruction quality across various ambient light sources, including indoor lighting and a bright work lamp. The reconstructions, shown for Hybrid $(n=255)$, are robust to ambient light, albeit with a drop in performance under the work lamp. These results can potentially be improved by judicious use of light redistribution schemes \cite{guptaStructuredLightSunlight2013,o2015homogeneous,matsudaMC3DMotionContrast2015} and exposure control.

\input{figures/fig_14_waving_cloth_banner}
\input{figures/fig_13_dark_objects}
\input{figures/fig_15_hand_gest}

Next, we consider low-albedo scenes by imaging objects covered by highly absorptive materials, such as 3M Black Matte \cite{3m_black_matte} and Acktar Velvet \cite{acktar}; the latter absorbs upto $99.9$\% of incident light. As \cref{fig:dark-objects} shows, Single-Photon SL can recover the 3D geometry of these dark objects, even when visually imperceptible. As a practical example, \cref{fig:banner-intro} shows the reconstruction of tire treads scanned at $40$ FPS.

\subsection{Dynamic Scenes with a High-Speed Projector}\label{sec:dynamic-scene-DLP}

For dynamic scenes, we used a projector based on the Texas Instruments DLP6500 DMD, capable of projecting binary images with a resolution of $1024\times 768$ pixels at 20 kHz. For simplicity, we operate the SPAD at the same speed as the projector. The projector uses a broadband white LED (SugarCUBE Ultra White LED) as the illumination source.

Figure \ref{fig:hand-gest} shows high-speed 3D imaging for a sequence of fast hand movements. A commercial 3D scanner (Kinect-v2 camera \cite{zhang2012microsoft}) operating at 30 FPS fails to recover the fingers of the rapidly moving hand, while Single-Photon SL continues to recover fine details. Finally, in \cref{fig:waving-cloth}, we reconstruct the deforming folds of a cloth as it is waved in front of the camera. For both sequences, we use Hybrid $(n=63)$ operated at $250$ FPS. These demonstrate the ability of proposed Single-Photon SL techniques to recover detailed 3D geometry of high-speed deformable objects.

In summary, our results on these challenging scenes---both static and dynamic---illustrates the practical capabilities of the Single-Photon SL modality and its ability to simultaneously achieve high speed, precision and robustness.

%% file: figures/fig_11_method_comp_unified.tex
\begin{figure*}[t]
    \centering
    \includegraphics[width=\textwidth]{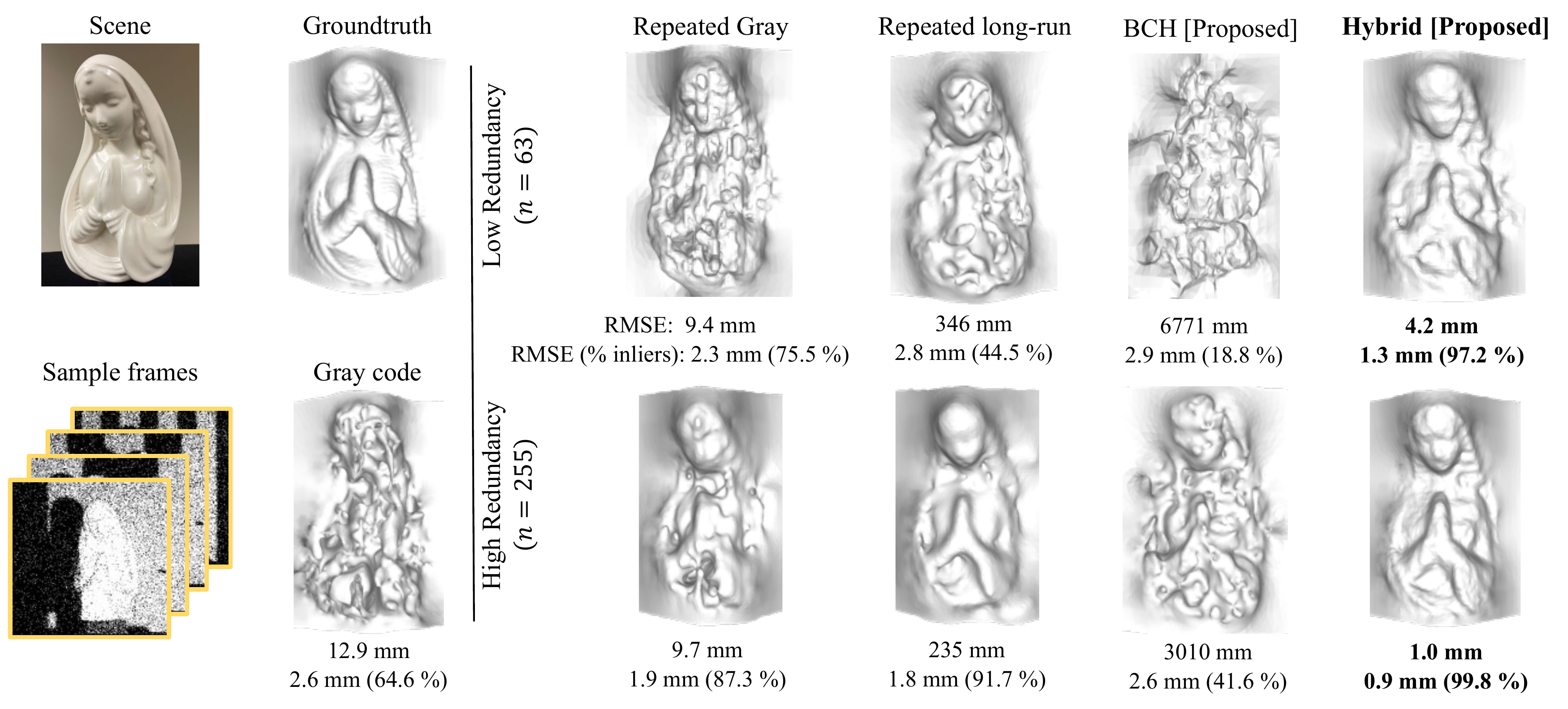}
    \vspace{-0.3in}
    \caption{\textbf{Strategy comparison for Single-Photon SL on porcelian bust.} \textit{(left)} We obtain ground truth by averaging the burst of $5120$ binary frames captured for each projected pattern. 
    To illustrate the challenge of photon noise, we include reconstruction using Gray code without repetition which has severe artifacts. \textit{(right)} Comparison between our proposed Hybrid strategy and other baseline methods across operating points $n = \{63,255\}$.  BCH codes having several high spatial-frequency frames, are easily distorted by short-range effects. We report three metrics: RMSE in estimated depth, percentage of inliers (absolute depth error $<5$mm), and RMSE among these inliers.}
    \vspace{-0.1in}
    \label{fig:method-comp-un}
\end{figure*}

%% file: figures/fig_12_ambient_light.tex
\begin{figure}[!ttt]
    \centering
    \includegraphics[width=0.85\columnwidth]{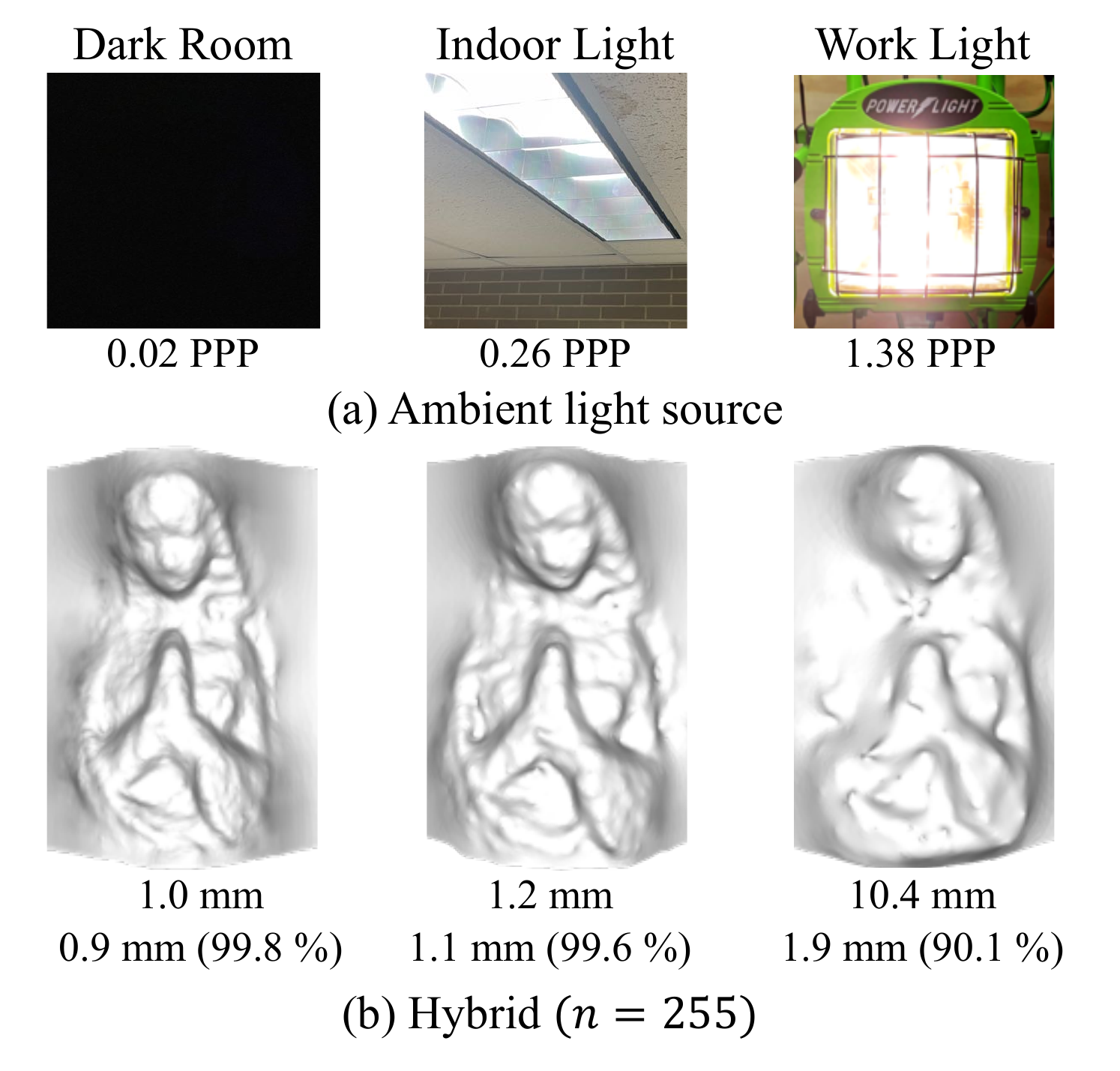}
    \vspace{-1em}
    \caption{\textbf{Effect of ambient light intensity.} Hybrid codes, shown here for $n=255$, are fairly robust to indoor ambient light and recover coarse shapes even under bright ambient light. We report the average Photons incident Per Pixel (PPP) during an exposure window, as a measure of ambient illumination.}
    \vspace{-0.15in}
    \label{fig:ambient-light}
\end{figure}

%% file: figures/fig_14_waving_cloth_banner.tex
\begin{figure*}[h]
    \centering
    \includegraphics[width=\textwidth]{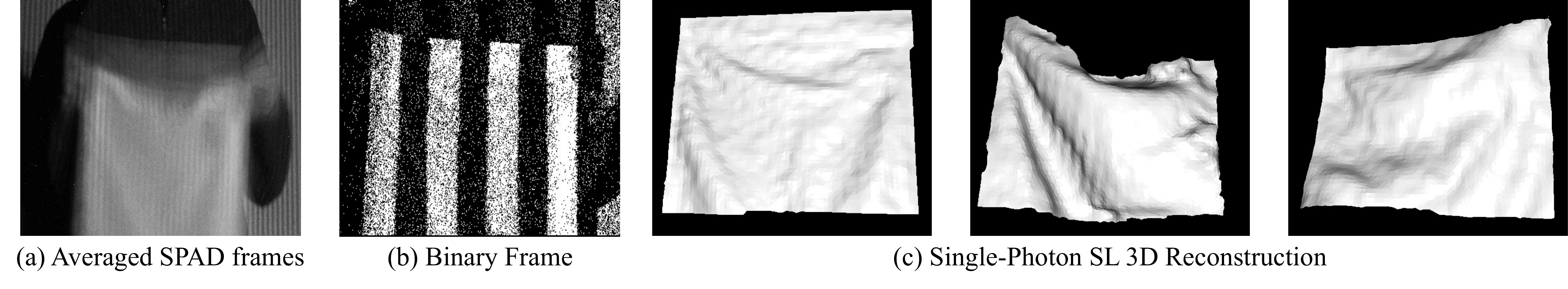}
    \vspace{-0.3in}
    \caption{\textbf{Non-rigid deforming object captured by Single-Photon SL} using Hybrid $(n=63)$ at $250$ FPS. We include (a) a reference image captured by the SPAD camera using a long integration time, (b) a single binary frame and (c) the reconstructed meshes clearly showing the folds of the cloth. Capturing a non-rigid object is particularly challenging---unless we operate at high speeds, excessive motion blur is induced. \textbf{We include a high-speed depth video of this sequence in the supplementary material.}}
    \label{fig:waving-cloth}
    \vspace{-0.25in}
\end{figure*}

%% file: figures/fig_13_dark_objects.tex
\begin{figure}[t]
    \centering
    \includegraphics[width=\columnwidth]{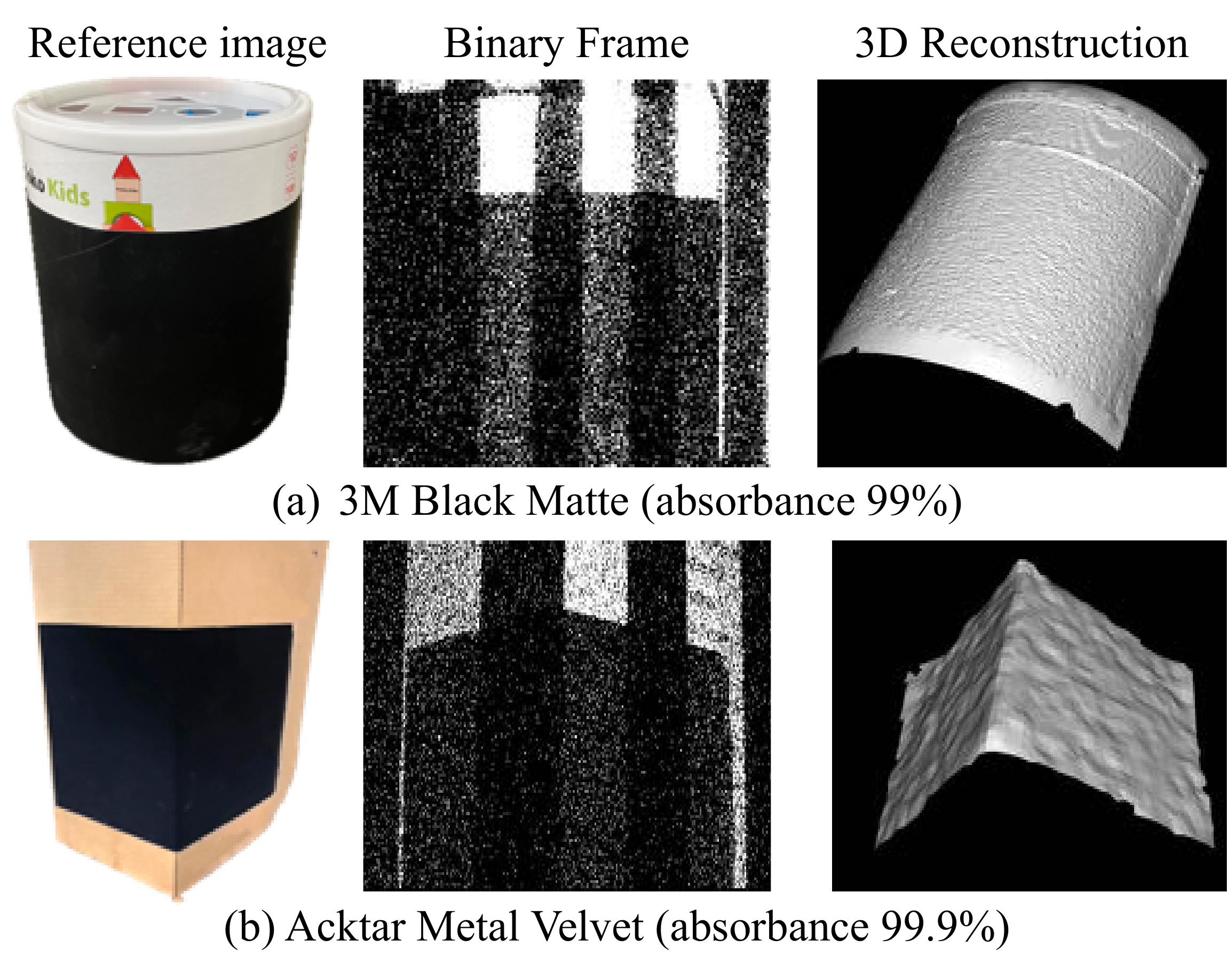}
    \vspace{-2em}
    \caption{\textbf{3D reconstruction in low-albedo scenes} of (a) a white cylinder covered by \textit{3M Black Matte} and (b) an inverted V-groove covered by \textit{Acktar Metal Velvet}; the materials used have extremely high absorbance of $99$\% and $99.9$\%, respectively. Both reconstructions are obtained using Hybrid $(n=255)$ at $40$ FPS.}
    \label{fig:dark-objects}
    \vspace{-0.15in}
\end{figure}

%% file: figures/fig_15_hand_gest.tex
\begin{figure}[t]
    \centering
    \includegraphics[width=\columnwidth]{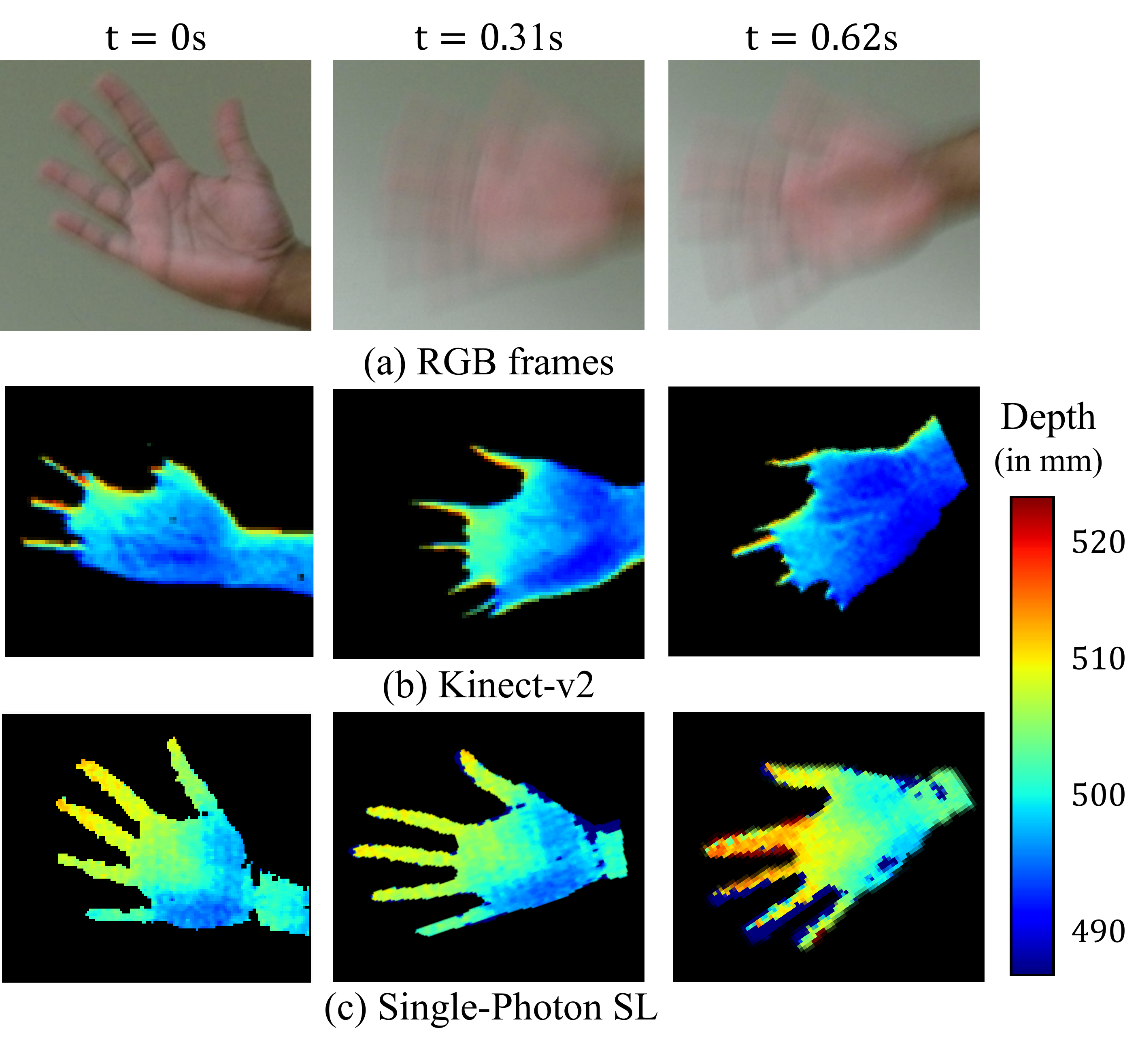}
    \vspace{-0.35in}
    \caption{\textbf{Comparison to Kinect-v2 on fast hand movements.} We use Hybrid codes $(n=63)$ at $250$ FPS to capture these rapid movements. Only the palm, and no fingers, is visible in the Kinect's depth maps. The motion blur distorted RGB frames, acquired by the Kinect at $30$ FPS, convey the speed involved.}
    \vspace{-0.2in}
    \label{fig:hand-gest}
\end{figure}

%% file: sections/conclusion.tex
This paper shows that many of the tradeoffs inherent to SL systems can be addressed via the use of SPAD sensors.
Single-Photon SL, the system that we propose, exploits the  single photon detection capabilities of SPAD sensors, along with its lack of read noise.
The proposed ideas are capable of detecting objects with high absorbance, and scenes with dramatic high-speed motion. The enabling techniques underlying these are a set of error correction codes, that are designed to be resilient to aberrations commonly present in SL systems.
As with many recent efforts in this space, Single-Photon SL provides yet another case study for the wider adoption of SPAD sensors in the imaging community.\smallskip

\noindent{\bf Optimal coding and decoding.} Despite their effectiveness, the proposed coding and decoding strategies are not provably optimal. 
Our MDD implementation is agnostic to the asymmetry of bit-flips, a defining feature of Single-Photon SL.
While maximum likelihood decoders remain to be constructed for the general asymmetric case, optimal decoders have been derived for edge cases \cite{ellingsen2005maximum}, e.g., when $\text{P}_\text{flip, bright}$\ggNice$\text{ P}_\text{flip, dark}$. 
Further, leveraging an optimization framework~\cite{Mirdeghan:2018} such as optical SGD \cite{Chen_2020_CVPR} can lead to improved coding and decoding schemes for Single-Photon SL.\smallskip

\noindent{\bf Handling ambient illumination.} In strong ambient illumination, bit-flips arising from $\text{ P}_\text{flip, dark}$ are predominant. An important next step is to exploit such asymmetry of bit-flips to design Single-Photon SL codes that are optimized for extreme ambient illumination. Finally, another promising research direction is to explore Single-Photon SL with complementary modalities such as light concentration~\cite{guptaStructuredLightSunlight2013,otoole2015} and epipolar structured light\cite{o2012primal,o20143d} to gain further robustness to extreme ambient and global illumination. 

%% file: supplementary_content.tex
\section{Deriving Expected Error} \label{supp_sec:expected_error}
\input{supplementary_sections/expected_error}
\clearpage

\section{Comparison to High-Speed Cameras} \label{supp_sec:high_speed_comp}
\input{supplementary_sections/high_speed_camera} 
\clearpage

\section{Code Look-Up-Tables} \label{supp_sec:code_LUTs}
\input{supplementary_sections/code_LUTs}
\clearpage

\section{Single-Photon SL Hardware Prototype} \label{supp_sec:hardware}
\input{supplementary_sections/hardware} 
\clearpage

\section{Benchmarking Minimum Distance Decoding Methods}  \label{supp_sec:mdd_bench}
\input{supplementary_sections/mdd_benchmark}
\clearpage

\section{More Results}  \label{supp_sec:more_results}
\input{supplementary_sections/more_results}
\clearpage

{\small

\input{supplementary.bbl}
}

%% file: supplementary_sections/expected_error.tex
In this supplementary note, we derive the decoding error probability of Gray codes and repetition codes subject to the bit-flip noise model (\cref{sec:noisemodels}) of Single-Photon SL. We provide an analytic expression for error probability in BCH codes for a given decoding function and for the particular case of a bounded-error decoder. Additionally, we derive an upper-bound on the average decoding error (in terms of absolute deviation) for binary shifted patterns. Finally, we outline the Monte-Carlo simulation procedure to compare various strategies---such as BCH and repetition strategies in \cref{fig:bch-repetition}, and Hybrid and repetition strategies in \cref{fig:hybrid-repetition}.

\subsection{Notation}

Let $\mathcal{M} = \{\mathbf{m} \;|\; \mathbf{m} \in \{0,1\}^L\}$ denote the set of $L$-bit binary messages, e.g., Gray codes, that represent columns of a projector. For simplicity, we shall assume that the projector has $2^L$ columns, but the following analysis extends mutatis mutandis to the more general case. Further, let  $\mathbf{1}_\mathbf{m}$ and  $\mathbf{0}_\mathbf{m}$
denote the number of $1$'s and $0$'s in $L$-bit message $\mathbf{m}$ respectively. 

\subsection{Gray Codes}
\label{supp_sec:gray_deriv}

When transmitting a Gray code $\mathbf{m}$ an error occurs if any of the bits are flipped due to photon-noise, since no error-correcting strategy is employed. Assuming $\mathbf{y}$ denotes the received (possibly corrupted) codeword when $\mathbf{m}$ is projected, the expected error probability over the set of messages $\mathcal{M}$ is given as

\begin{align}
    \Prob{\text{error}} &= \sum_{\mathbf{m} \in \mathcal{M}} \Prob{\mathbf{m}} \Prob{\cond{\text{error}}{\mathbf{m}}} \nonumber \\
    &= \frac{1}{2^L} \sum_{\mathbf{m} \in \mathcal{M}} \Prob{\cond{\mathbf{y} \neq \mathbf{m}}{\mathbf{m}}} \tag*{(Each message is equally likely to be transmitted)}\\
    &= \frac{1}{2^L} \sum_{\mathbf{m} \in \mathcal{M}} 1 - \Prob{\cond{\mathbf{y} = \mathbf{m}}{\mathbf{m}}} \nonumber  \\
    &= \frac{1}{2^L} \sum_{\mathbf{m} \in \mathcal{M}} 1 - \prod_{i=1}^L \Prob{\mathbf{y}_i = \mathbf{m}_i} \nonumber \\
    &= 1 - \frac{1}{2^L} \sum_{\mathbf{m} \in \mathcal{M}} \Prob{\cond{\mathbf{y}_i = 1}{\mathbf{m}_i=1}}^{\mathbf{1}_\mathbf{m}} \Prob{\cond{\mathbf{y}_i = 0}{\mathbf{m}_i=0}}^{\mathbf{0}_\mathbf{m}} \nonumber  \\
    &= 1 - \frac{1}{2^L} \sum_{i=0}^L \binom{L}{i} \Prob{\cond{\mathbf{y}_i = 1}{\mathbf{m}_i=1}}^{i} \Prob{\cond{\mathbf{y}_i = 0}{\mathbf{m}_i=0}}^{L-i} \nonumber  \\
    &= 1 - \frac{1}{2^L} \left( \Prob{\cond{\mathbf{y}_i = 1}{\mathbf{m}_i=1}} + \Prob{\cond{\mathbf{y}_i = 0}{\mathbf{m}_i=0}}\right)^L \nonumber \\
    &= 1 -  \left( 1 -
    \frac{\left(\text{P}_\text{flip, bright}+\text{P}_\text{flip, dark}) 
    \right)}{2}
    \right)^L \label{eq:gray-deriv}
\end{align}

\noindent where, we make use of \cref{eq:flipbright} and \cref{eq:flipdark} in the last step. Note that the decoding error is zero when $\text{P}_\text{flip, bright}, \text{P}_\text{flip, dark} \to 0$. 

\clearpage
\subsection{Repeated Gray Codes}

Assume we repeat $\mathbf{m}$, the $L$-bit message, $r$ times and decode it with a majority vote. Let $\mathbf{x}$ denote the received codeword and $\mathbf{y}$ the decoded message by using a majority vote. Then each bit $y_i$ is decoded incorrectly if more than or equal $\ceil{\frac{r-1}{2}}$ bit-flips occur, with probability

\begin{align}
    \Prob{\cond{\mathbf{y}_i \neq \mathbf{m}_i}{\mathbf{m}_i=1}} &= \prod_{j=\ceil{\frac{r-1}{2}}}^r \binom{r}{j} \text{P}_\text{flip, bright}^j (1- \text{P}_\text{flip, bright})^{r-j} \nonumber \\
    &= g_r(\text{P}_\text{flip, bright})
\end{align}

where $g_r(p) = \prod_{j=\floor{\frac{r-1}{2}}}^r p^r (1-p)^{r-j}$. Similarly, we can show that 

\begin{align}
    \Prob{\cond{\mathbf{y}_i \neq \mathbf{m}_i}{\mathbf{m}_i=0}} 
    &= g_r(\text{P}_\text{flip, dark}) \nonumber
\end{align}

With this, we can evaluate decoding error probability for the repetition strategy by extending \cref{eq:gray-deriv} as

\begin{align}
    \Prob{\text{error}} &= 1 -  \left( 1 -
    \left (\frac{g_r\left(\text{P}_\text{flip, bright} \right)
    +
    g_r\left(\text{P}_\text{flip, dark} \right )
    }{2}\right)
    \right)^L
\end{align}

\noindent Note that, $g_r(p) \geq p\; \forall \; p \in [0,1]$, so repetition always improves decoding reliability when $\text{P}_\text{flip, dark},  \text{P}_\text{flip, bright} < 0.5$.

\clearpage
\subsection{BCH Codes}

\paragraph{Construction.} A $\BCH{n}{k}{d}: \{0,1\}^k \to \{0, 1\}^n$ encoder takes input messages of length $k$ and produces output codewords of length $n$ that are at least $d$-bits apart. In this paper, we consider primitive, narrow-sense BCH codes over the finite field $\text{GF}(2)$. We present a brief summary of their construction here and refer readers to Roth \cite{roth2006introduction} for a more detailed explanation.

\begin{enumerate}
    \item Pick $\alpha$ a primitive element of $\text{GF}(2^n)$.
    \item Compute the generator polynomial $g(x) = \text{lcm}\left(q_1(x),...,q_{d-1}(x)\right)$, where $q_i(x)$ are minimal polynomials of $\alpha^i$ with coefficients in $\text{GF}(2)$.
    \item Each codeword is obtained by multiplying a polynomial representing the message $p(x)$ and the generator polynomial $g(x)$. 
    \item For systematic encoding, we set $p(x) = m(x)x^{n-k} - r(x)$, where $m(x)$ is the polynomial with the symbols of message $\mathbf{m}$ as coeffecients and $r(x) = m(x)x^{n-k} \text{ mod } g(x)$.
\end{enumerate}

\noindent From the construction, it can be seen that BCH codes are a subset of Reed-Solomon codes over the finite field $\text{GF}(2^n)$ \cite{guruswami2012essential}. Specifically, $\mathcal{C}_\BCH{n}{k}{d} = \mathcal{C}_{\text{RS}(n,k,d)} \cap \{0,1\}^n$, where $\text{RS}(n,k,d)$ represents the Reed-Solomon encoder with parameters $(n,k,d)$ as before.

\paragraph{Decoding Error.} Let $\mathbf{c}$ denote a codeword corresponding to BCH encoded message $\mathbf{m}$. As before, assume we receive (possibly corrupted) bits $\mathbf{x}$ which we decode to obtain $\mathbf{y}$. Depending on whether decoder gives up beyond the worst-case error limit (e.g. Berlekamp Massey \cite{berlekamp_massey}) or not (e.g. minimum distance decoding), we can derive two expressions. For decoders that work up to the worst-case error limit of $\floor{\frac{d - 1}{2}}$ bit-flips, 

\begin{align}
    \Prob{\mathbf{y} \neq \mathbf{m}} &= \Prob{d_H(\mathbf{x}, \mathbf{c}) > \frac{d - 1}{2}} \tag*{(where $d_H$ is the Hamming distance)} \\
    &= \Prob{\text{\#bit-flips bright + \#bit-flips dark} > \frac{d - 1}{2}} \nonumber \\
    &= \sum_{j = 0}^{\frac{d - 1}{2}} \Prob{\text{\#bit-flips bright} = j}\Prob{\text{\#bit-flips dark} > \frac{d - 1}{2} - j} \nonumber \\
    &= \sum_{j = 0}^{\frac{d - 1}{2}} \sum_{k=\frac{d - 1}{2} - j}^{\mathbf{1}_c} \binom{\mathbf{0}_c}{j} \binom{\mathbf{1}_c}{k} \text{P}_\text{flip, dark}^j (1 - \text{P}_\text{flip, dark})^{\mathbf{0}_c - j} \nonumber\\
    &\quad \text{P}_\text{flip, bright}^k (1 - \text{P}_\text{flip, bright})^{\mathbf{1}_c - k}
    \label{eq:bch-bounded-error}
\end{align}

Whereas, if we use a minimum distance decoder,

\begin{align}
    \Prob{\mathbf{y} \neq \mathbf{m}} &= \Prob{\min_{\mathbf{z} \in \mathcal{C}} d_H(x,z) \neq c} \label{eq:bch-mdd-error} \\
    & \tag*{, where $\mathcal{C} = \Set{\mathbf{c}}{\mathbf{c} = \mathcal{E}_{\BCH{n}{k}{d}}(\mathbf{m}) \, \forall \, \mathbf{m} \in \mathcal{M}}$}
\end{align}

Evaluating this probability expression in closed-form is not straightforward, but can be done by enumerating each code-word in an exhaustive manner. For this reason, we opt to use Monte-Carlo simulations when comparing BCH and repetition strategies.

\clearpage
\subsection{Binary Shifting}
\label{supp_sec:binary-shift}

Decoding binary shifted patterns is achieved by using a matched filter, where the template corresponds to a pattern with a burst of $2^{L_\text{shift}}$ ones followed by $2^{L_\text{shift}}$ zeros. In this section, we present the error analysis for binary shifting with $L_\text{shift}=3$, corresponding to a temporal pattern length of $2^{L_\text{shift} + 1}= 16$ frames. Without loss of generality, we consider the temporal sequence representing pixel location $0$---comprising of 8 ones followed by 8 zeros. The expected error of other circularly shifted sequences are identical to this. Let $s_t$ denote the value of the received signal (possibly corrupted) at the $t$-th time instant ($0 \leq t \leq 15$). Then, the probability of the absolute decoding error being exactly 1 pixel can be derived as

\begin{align}
    \Prob{|\text{error}|=1} &= \Prob{\{\sum_{t=1}^8 s_t \geq \sum_{t=j}^{7+j} s_t \;\forall\; j\} \bigcup \{\sum_{t=-1}^6 s_t \geq \sum_{i=j}^{7+j} s_i \;\forall\; j\}} \nonumber\\
    &\leq 2\Prob{\{\sum_{t=1}^8 s_t \geq \sum_{t=0}^{7} s_t \}} \tag*{(Union bound, equally likely events)}\\
    &= 2\Prob{s_8 \geq s_0} \nonumber\\
    &= 2 \Prob{X \geq Y} \tag*{($X \sim \text{Ber}(\text{P}_\text{flip, dark}),\; Y \sim \text{Ber}(1 - \text{P}_\text{flip, bright})$)}\\ \\
    &= 2\left(\text{P}_\text{flip, dark} + (1-\text{P}_\text{flip, dark})\text{P}_\text{flip, bright}\right)
\end{align}

where, $\text{Ber}(p)$ denotes the Bernoulli probability distribution. In the above expression, we consider time instances modulo 16, i.e., $t=-1$ is interpreted as $t=15$. Next, the probability of the absolute decoding error being exactly 2 pixels can computed as

\begin{align}
    \Prob{|\text{error}|=2} &= \Prob{\{\sum_{t=2}^9 s_t \geq \sum_{t=j}^{7+j} s_t \;\forall\; j\} \bigcup \{\sum_{t=-2}^5 s_t \geq \sum_{t=j}^{7+j} s_t \;\forall\; j\}} \nonumber\\
    &\leq 2\Prob{\{\sum_{t=2}^9 s_t \geq \sum_{t=0}^{7} s_t \}} \tag*{(Union bound)} \nonumber\\
    &= 2\Prob{s_8 + s_9 \geq s_0 + s_1}\\
    &= 2 \Prob{X \geq Y} \tag*{($X \sim \text{Bin}(2, \text{P}_\text{flip, dark}),\; Y \sim \text{Bin}(2, 1 - \text{P}_\text{flip, bright})$)}
\end{align}

where, $\text{Bin}(n,p)$ denotes the binomial probability distribution. Similarly, we can upper bound the probability of the absolute decoding error being $1 \leq r \leq 15$ pixels as $2\Prob{X_r \geq Y_r}$, where $X_r \sim \text{Bin}(r, \text{P}_\text{flip, dark}),\; Y_r \sim \text{Bin}(r, 1 - \text{P}_\text{flip, bright})$. With this, the expected absolute decoding error is given by:

\begin{align}
    \mathbb{E}[|\text{error}|] &= \sum_{r=0}^{15} r \Prob{|\text{error}| = r} \nonumber\\
    &\leq \sum_{r=1}^{15} 2 r \Prob{X_r \geq Y_r}
\end{align}

Using this expression, the upper bound on the expected decoding error for the dark room condition (using flux values from \cref{fig:prob}\textcolor{red}{(a)}) is $1.2$ pixels.

\clearpage

\subsection{Monte-Carlo Simulation Procedure}

\begin{algorithm}[h]
\caption{Monte-Carlo Simulation Procedure}\label{alg:monte-carlo}
\begin{algorithmic}
\Require {Bit-flips probabilities $\text{P}_\text{flip, bright}, \text{P}_\text{flip, dark}$ \\
Message set $\mathcal{M}$ with corresponding codeword set $\mathcal{C}$\\
error metric $\mathcal{L}$\\
Monte-Carlo iterations $n_\text{iter}$ \;
\;}
\Procedure{Monte-Carlo-Simulation}{$\text{P}_\text{flip, bright}$, $\text{P}_\text{flip, dark}$, $\mathcal{M}$, $\mathcal{C}$, $\mathcal{L}$ $n_\text{iter}$}
\State $\text{error} \gets 0$
\For{$\mathbf{m} \in \mathcal{M}$}
\State Pick corresponding $\mathbf{c} \in \mathcal{M}$
\State $\text{error}_\mathbf{m} \gets 0$
\For{$1 \leq i \leq n_\text{iter}$}
\State Randomly flip $\mathbf{1}_\mathbf{c}$ and $\mathbf{0}_\mathbf{c}$ with probability $\text{P}_\text{flip, bright}$ and $\text{P}_\text{flip, dark}$ respectively to obtain $\mathbf{x}$
\State Decode $\mathbf{x}$ to obtain $\mathbf{y}$
\State Compute $\mathcal{L}(\mathbf{m}, \mathbf{y})$
\State $\text{error}_\mathbf{m} = \text{error}_\mathbf{m} +  \mathcal{L}(\mathbf{m}, \mathbf{y})$ 
\EndFor
\State $\text{error}_\mathbf{m} = \text{error}_\mathbf{m} / n_\text{iter}$
\State $\text{error} = \text{error} + \text{error}_\mathbf{m}$
\EndFor
\State $\text{error} = \text{error} / |\mathcal{M}|$ \\
\Return $\text{error}$
\EndProcedure
\end{algorithmic}
\end{algorithm}

We describe the simulation procedure in \cref{alg:monte-carlo}. The error metrics we consider are exact error ($\sum_{i=1}^N \frac{1}{N} \mathbb{I}(x_i,y_i)$, where $\mathbb{I}$ denotes the indicator function) and root mean square error (RMSE, $\sqrt{\frac{1}{N} \sum_{i=1}^N (x_i - y_i)^2}$). \Cref{fig:bch-repetition} and \Cref{fig:hybrid-repetition} both use $n_\text{iter}=100$ (Monte-Carlo iterations).

\clearpage

\subsection{Empirical Comparison of BCH and Repetition Strategies at Various $\Phi_a$}

\input{supplementary_sections/figures/bch_repetition}

%% file: supplementary_sections/figures/bch_repetition.tex
\begin{figure}[h]
    \centering
    \includegraphics[width=\textwidth]{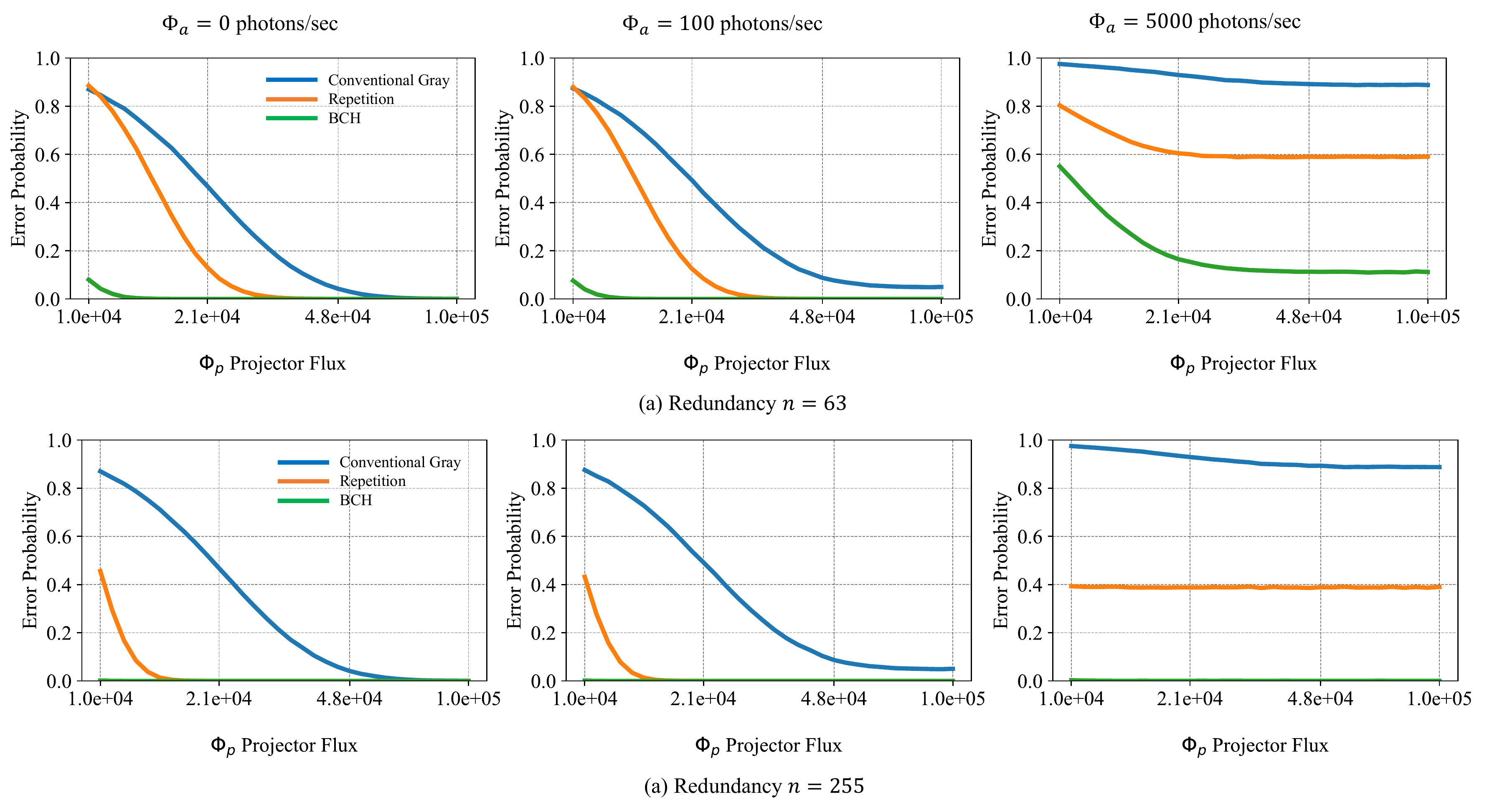}
    \vspace{-0.2in}
    \caption{\textbf{Monte-Carlo evaluation of BCH and repetition strategies,} plotted at various ambient flux $\Phi_a$. $\Phi_a=0$ photons/sec (\textit{left column}) represents a dark room, while $\Phi_a=100$ photons/sec and $\Phi_a=5000$ photons/sec \textit{(middle and right columns)} represent indoor illumination and bright intensity matching the projector's flux respectively. Across ambient illumination levels and redundancy factors, BCH codes outperform repetition codes. Particularly at $n=255$, BCH codes result in almost zero error everywhere.}
    \label{fig:supp-bch-rep}
\end{figure}

%% file: supplementary_sections/high_speed_camera.tex
Read-noise of existing high-speed cameras is quite large and worsens with increasing read-out rates \cite{readout_noise}.
For instance, the Phantom v2640 \cite{phantom_v2640}
has a read noise of 18.8e- when capturing $640 \times 480$ frames at 28 kHz. This leads to an order of magnitude lower SNR than SPADs (\cref{tab:supp_SNR_values}), and hence extremely noisy, practically unusable, reconstructions (\cref{fig:supp-hand-speed-comp}). This is because high-speed cameras must capture sufficient photons in each frame to beat the high read noise floor, which is not possible especially in high-speed motion scenarios considered in this paper.

\begin{table}[h]
    \centering
    \begin{tabular}{@{}cccc@{}}\toprule
        Device & Dark room & Indoor lamp & Spot lamp \\
        \midrule
        Phantom-v2640 & 0.07 & 0.08 & 0.13 \\
        SwissSPAD & 0.81 & 0.84 & 1.10 \\
        \bottomrule
    \end{tabular}
    \caption{\textbf{SNR of a high-speed camera (Phantom-v2640) vs a SPAD camera (this work),} when observing a single bright pattern at 20kHz across various ambient conditions. No new experiment was conducted---incident flux data is obtained from Fig 3.}
    \label{tab:supp_SNR_values}
\end{table}

\input{supplementary_sections/figures/high_speed_sim}

%% file: supplementary_sections/figures/high_speed_sim.tex
\begin{figure}[h]
    \centering
    \includegraphics[width=0.7\columnwidth]{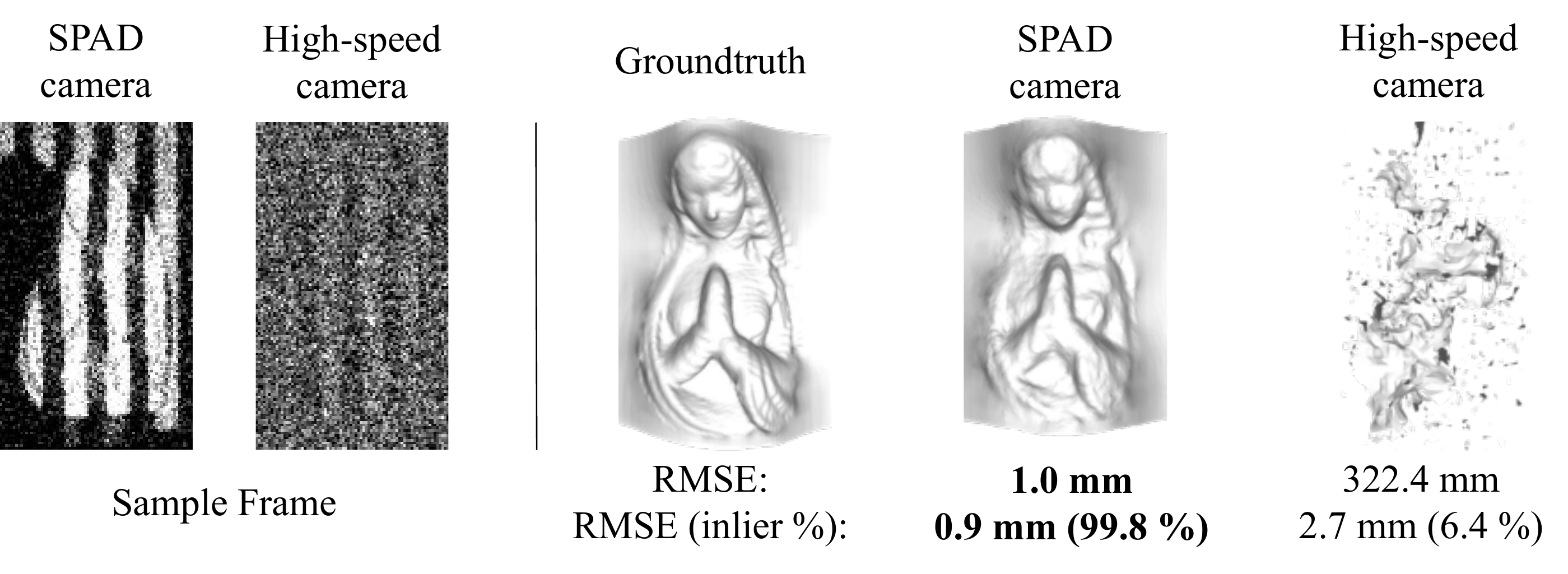}
    \caption{\textbf{3D reconstruction from the Phantom-v2640 vs the SPAD camera.} \textit{(left)} We simulate high-speed capture using photon arrival rates.  \textit{(right)} Read-noise dominated frames of the high-speed camera lead to poor reconstructions. For a fair comparison, we use Hybrid-255 and Hybrid-127 with complementary frames as the coding scheme for the SPAD camera and the high-speed camera respectively---requiring similar acquisition times.}
    \label{fig:supp-hand-speed-comp}
\end{figure}

%% file: supplementary_sections/code_LUTs.tex
In this note, we describe the coding strategies developed such as BCH and Hybrid encoding, by means of their code look-up-tables (LUTs)---which depict the projected patterns row-wise across time.

\subsection{\BCH{31}{11}{11} Encoding of a 10-bit Conventional Gray Message}
\input{supplementary_sections/figures/bch_31_gray}
\clearpage

\subsection{\BCH{31}{11}{11} Encoding of a 10-bit Long-run Gray Message}
\input{supplementary_sections/figures/bch_31_long_run}
\clearpage

\subsection{Hybrid $(n=31)$ Encoding of a 10-bit Gray Message}
\input{supplementary_sections/figures/hybrid_31_gray}

%% file: supplementary_sections/figures/bch_31_gray.tex
\begin{figure}[htp]
    \centering
    \includegraphics[width=\textwidth]{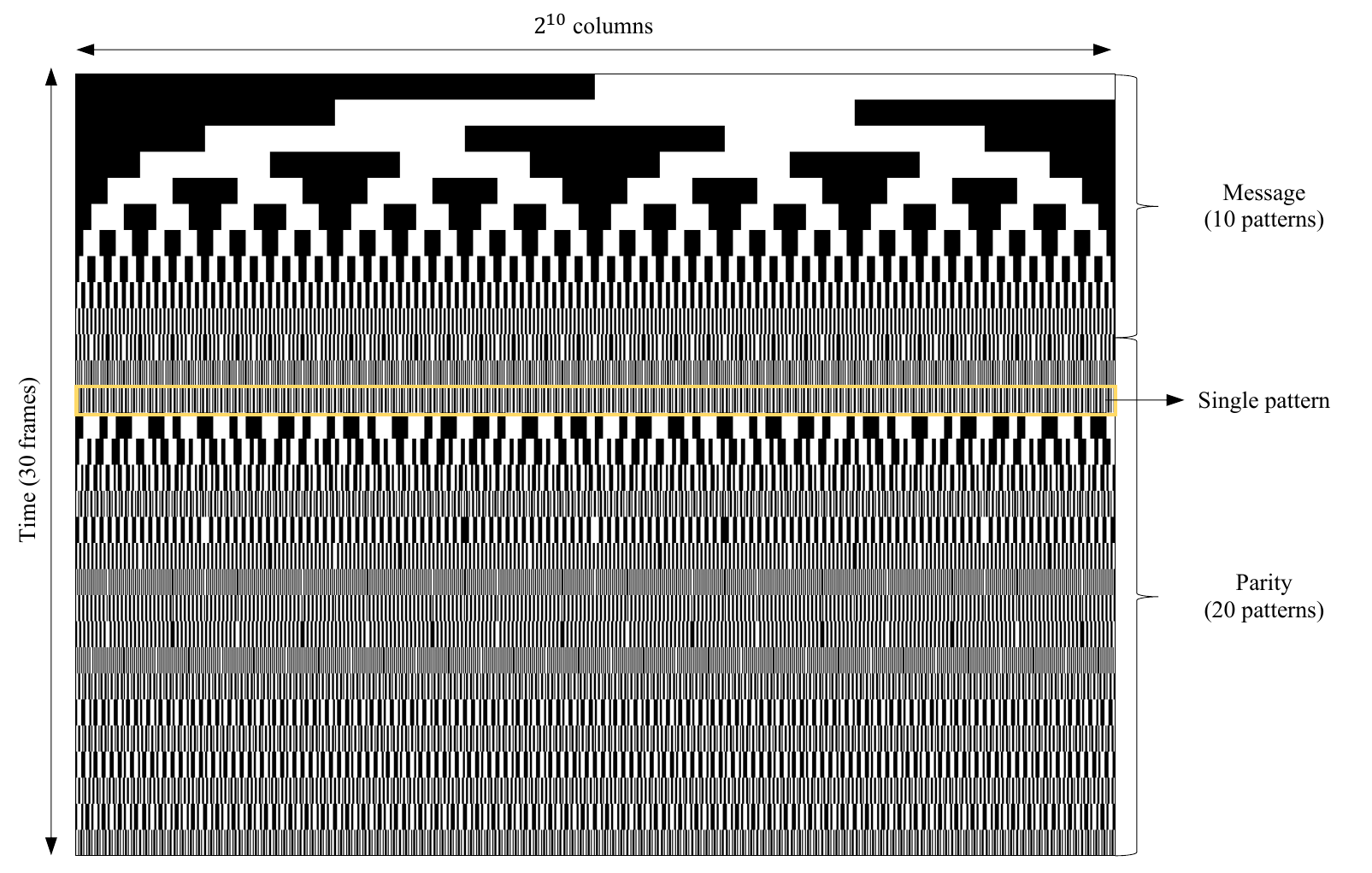}
    \caption{\textbf{Code Look-Up-Table describing the \BCH{31}{11}{11} encoding of a 10-bit Conventional Gray message.} Following systematic encoding, the code LUT upto the first 10 frames is identical to a Gray Code LUT. Most of the parity frames feature high spatial-frequency, which is easily distorted by short-range effects in a SL system (\cref{sec:practical-cons}). Zoom-in to see high-frequency details.}
    \label{fig:supp-bch-31-gray}
\end{figure}

%% file: supplementary_sections/figures/bch_31_long_run.tex
\begin{figure}[htp]
    \centering
    \includegraphics[width=\textwidth]{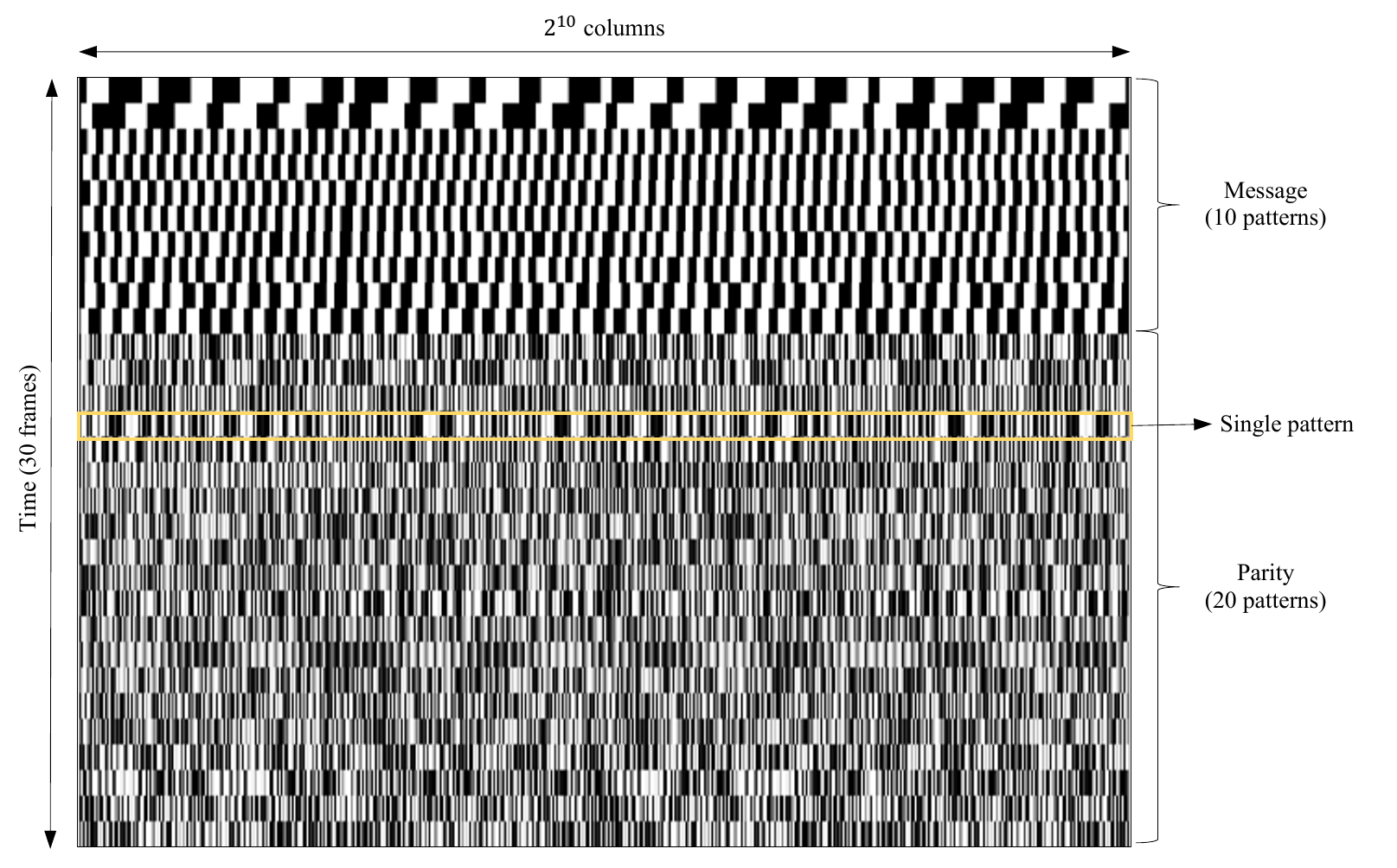}
    \caption{\textbf{Code Look-Up-Table describing the \BCH{31}{11}{11} encoding of a 10-bit Long-run Gray message.} Owing to the maximal minimum stripe-widths of Long-run Gray codes, the first 10 frames of the code LUT are robust to short-range distortions such as projection defocus and/or resolution mismatch. However, the parity frames continue to comprise of high-spatial frequency patterns and are thus, easily distorted. Zoom-in to see high-frequency details.}
    \label{fig:supp-bch-31-long-run}
\end{figure}

%% file: supplementary_sections/figures/hybrid_31_gray.tex
\begin{figure}[htp]
    \centering
    \includegraphics[width=\textwidth]{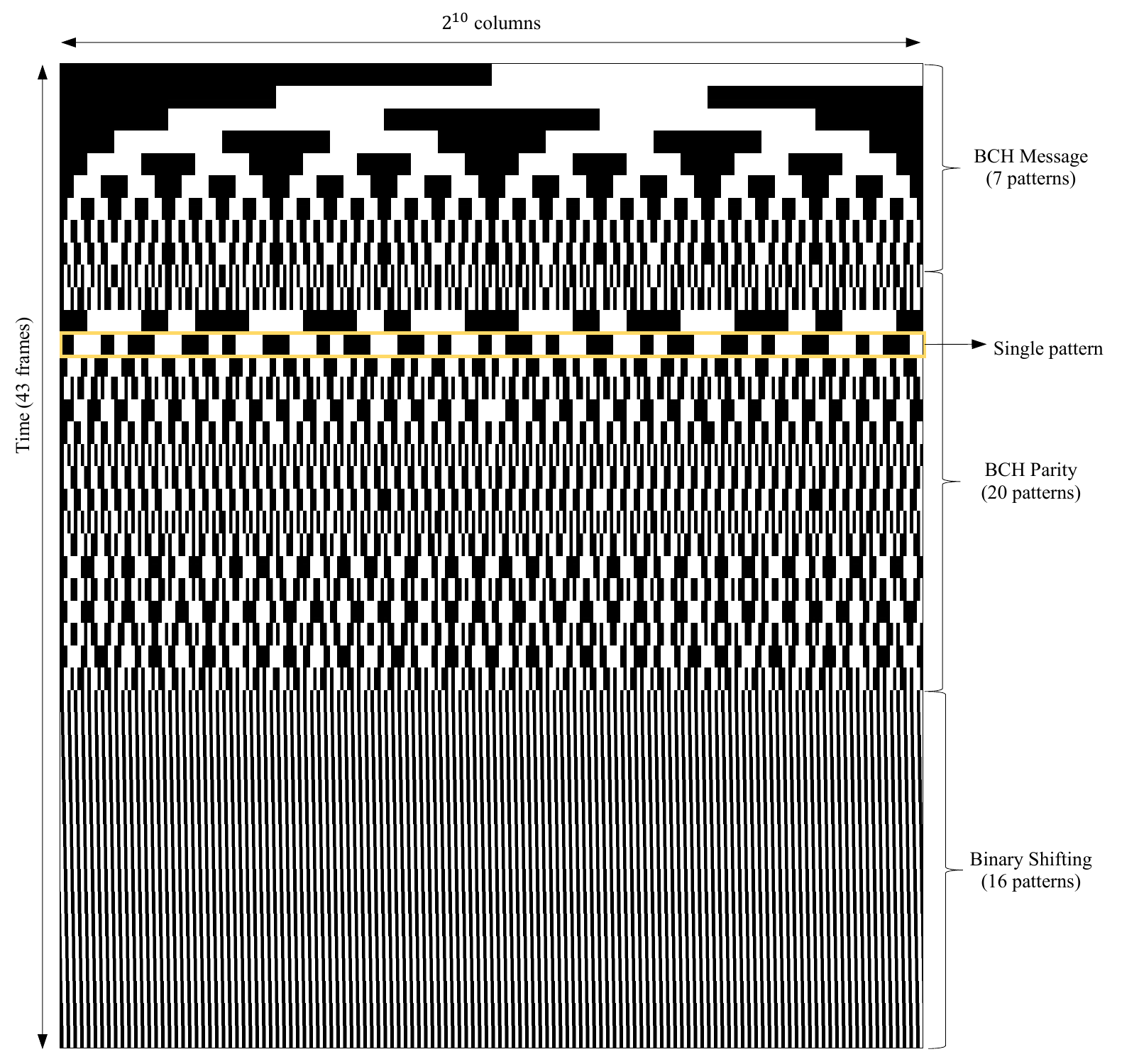}
    \caption{\textbf{Code Look-Up-Table describing the Hybrid$(n=31)$ encoding of a 10-bit Conventional Gray message.} We encode $L_\text{BCH}=7$ MSBs using the \BCH{31}{11}{11} encoder and $L_\text{shift}=3$ LSBs using binary shifting. Unlike the code LUTs of BCH encoded patterns (\cref{fig:supp-bch-31-gray} and \cref{fig:supp-bch-31-long-run} which comprise of many high spatial-frequency patterns (typically featuring a minimum stripe-width of 1 or 2 pixels), the minimum stripe-width of these Hybrid $(n=31)$ codes are at least 8 pixels by construction. This results in Hybrid codes offering robustness to both bit-flips arising from photon noise and other short-range effects arising from a combination of projector defocus and camera-projector resolution mismatch.}
    \label{fig:supp-hybrid-31-gray}
\end{figure}

%% file: supplementary_sections/hardware.tex
\input{supplementary_sections/figures/single_photon_sl_setup}

Supp. Fig. \ref{fig:supp-setup} shows the various components of our Single-Photon SL prototype. The resolution of the SPAD array is $512 \times 256$ pixels, while the DLP projector has a resolution of $1024 \times 768$ pixels. We used a baseline of 14 cm between the camera and projector---with the intention of maximizing the field of view of the camera occupied by the objects of interest.

\subsection{Calibration Procedure}

Our calibration procedure is similar to Zhang et al. \cite{zhang2006novel}, where the projector is treated as an inverse camera. Specifically, we capture correspondence maps of a chessboard, repeated across $30$ different positions. We then perform a calibration step mimicking a stereo setup based on the detected corners of the chessboard and their row and column correspondences. \smallskip

To obtain correspondences, we use projected patterns based on Gray codes and binary shifting---similar to our proposed Hybrid strategy, but with no added redundancy---to encode both column and row indices. This provides us precise row and column correspondence while simultaneously dealing with short-range effects arising from projector defocus / resolution-mismatch. To suppress the effects of photon-noise, we capture $5120$ SPAD frames for each projected pattern---which are subsequently averaged and compared against acquired complementary frames to produce binary outputs.

%% file: supplementary_sections/figures/single_photon_sl_setup.tex
\begin{figure}[htp]
    \centering
    \includegraphics[width=\textwidth]{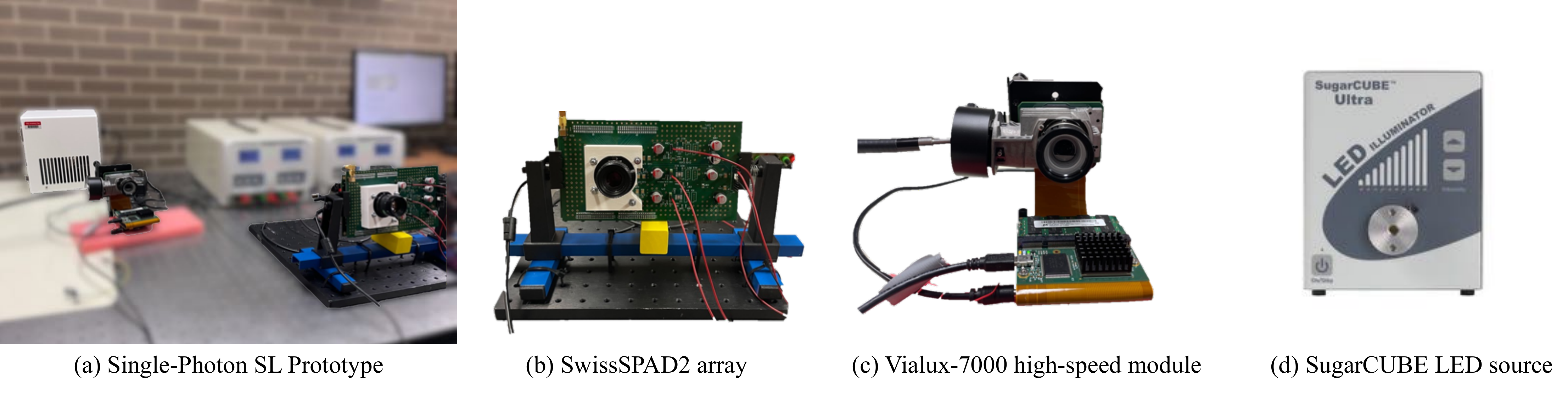}
    \caption{\textbf{Single-Photon SL hardware prototype.} The setup consists of a SwissSPAD2 array \cite{ulku512512SPAD2019} and a Vialux-7000 development kit which is based on the Texas Instruments DLP6500 DMD. For an illumination source we use a SugarCUBE Ultra, a broadband white LED, featuring a light output of $4000$ lumens.}
    \label{fig:supp-setup}
\end{figure}

%% file: supplementary_sections/mdd_benchmark.tex
Supp. Fig. \ref{fig:supp-bench} shows that Minimum Distance Decoding (MDD) can be significantly faster than algebraic decoding in Single-Photon SL. Supp. Fig. \ref{fig:supp-faiss} further shows that the most performant MDD method---FAISS using a GPU runtime---also scales reasonably to larger sized arrays including 1MPixel and 4MPixel sensors.

\input{supplementary_sections/figures/minimum_distance_bench}

\input{supplementary_sections/figures/faiss_scaling}

%% file: supplementary_sections/figures/minimum_distance_bench.tex
\begin{figure}[htp]
    \centering
    \includegraphics[width=0.5\columnwidth]{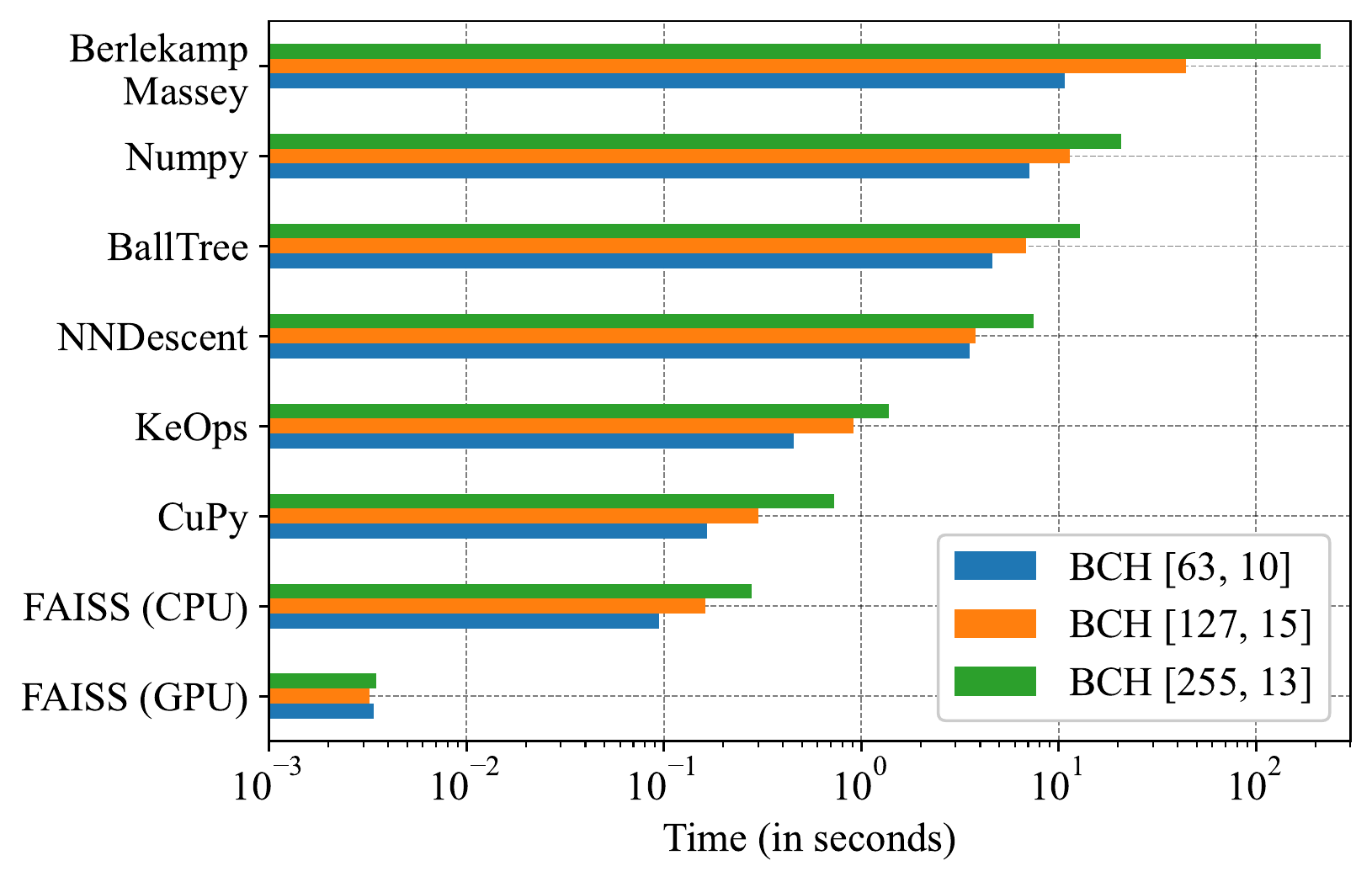}
    \caption{\textbf{Benchmark of various Minimum Distance Decoding (MDD) methods.} We query a $512 \times 256$ array of  codewords each encoding a $10$-bit message. We compare an algebraic method (Berkelamp-Massey \cite{berlekamp_massey}) to several MDD methods including those using exhaustive search (Numpy \cite{harris2020array}, CuPy \cite{Okuta2017CuPyA}), graph-based search (BallTree \cite{balltree_boytsov2013engineering}, NNDescent \cite{nndescent_dong2011efficient}) symbolic-tensors (KeOps \cite{keops_JMLR:v22:20-275}) and batched queries (FAISS \cite{faiss_johnson2019billion}). Benchmarks reported across various BCH encoders using a Intel-i7 8700K CPU and a NVIDIA 1080Ti GPU.}
    \label{fig:supp-bench}
\end{figure}

%% file: supplementary_sections/figures/faiss_scaling.tex
\begin{figure}[htp]
    \centering
    \includegraphics[width=\textwidth]{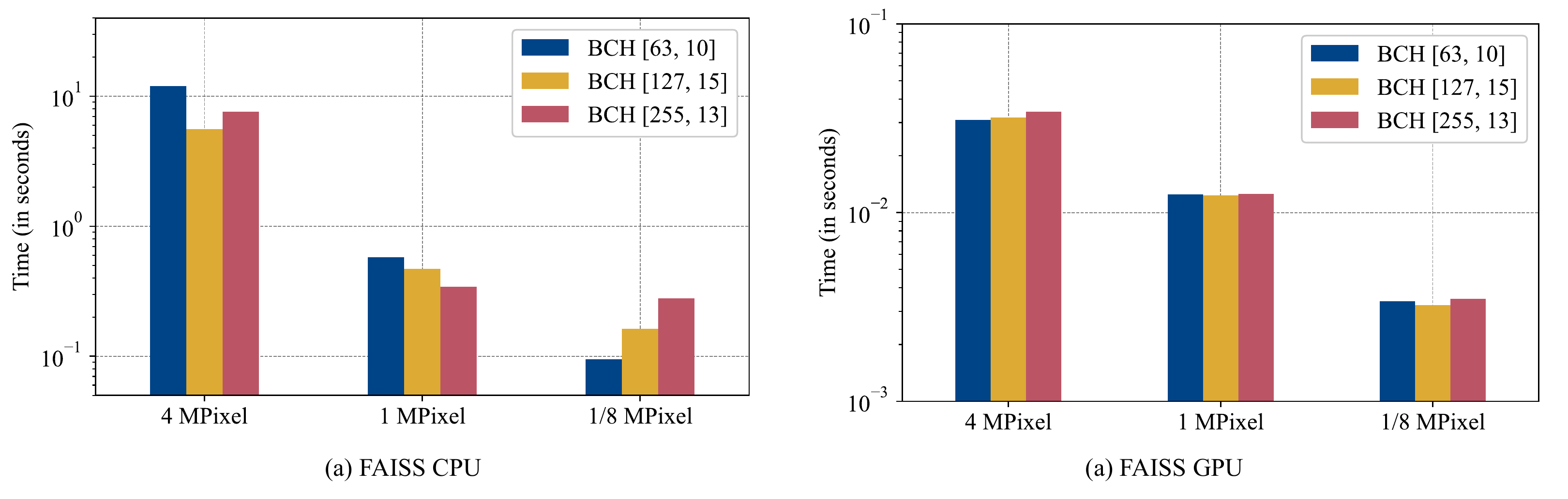}
    \caption{\textbf{MDD methods scale to larger resolution arrays.} Our implementation based on FAISS requires about $4$s and $15$s on a CPU (Intel-i7); $12$ ms and $30$ ms for one and four megapixels respectively on a GPU device (GTX 1080Ti).}
    \label{fig:supp-faiss}
\end{figure}

%% file: supplementary_sections/more_results.tex
\subsection{Low-albedo Objects}

Supp. Fig. \ref{fig:supp-dark} includes additional results on low-albedo objects; demonstrating the capabilities of Single-Photon SL in 3D imaging commonplace dark objects---such as the teflon pan, coffee mug and car headrest.
\input{supplementary_sections/figures/dark_objects}

\clearpage

\subsection{Dynamic Motion Sequences}

We include further results on dynamic scenes in Supp. Fig. \ref{fig:supp-dyn}. The curtain took an estimated $70$ milliseconds to fall, while the Jack-in-the-box toy took around $40$ milliseconds to spring up.

\paragraph{Going faster with pipelining.} 

While we have decoded disjoint bursts of acquired frames to output depth-maps in \cref{fig:waving-cloth} and \cref{fig:hand-gest}, we can also use overlapping sequences. Assuming $N$ patterns are projected, we can recover depth across time by decoding sequences $\{I_i\}_{i=1}^N, \{I_i\}_{i=N+1}^{2N}, \{I_i\}_{i=2N+1}^{3N},...$, or by considering sequences $\{I_i\}_{i=1}^N, \{I_i\}_{i=s+1}^{N+s}, \{I_i\}_{i=s+2}^{N+s+1},...$ i.e., using a sliding window of $N$ patterns with stride $0 < s < N$. This strided or pipelined approach contains more temporal information than disjoint bursts. We utilize pipelining based frame-interpolation for the jack-in-the-box sequence to produce a high-speed 3D reconstruction video at $2000$ FPS. 

\paragraph{High-speed depth videos} for the deforming cloth sequence and the Jack-in-the-box toy can be found on the \href{https://wisionlab.cs.wisc.edu/project/single-photon-sl/}{project webpage}. By using pipelining-based decoding, we output videos at $1000$ FPS for the waving cloth and $2000$ FPS for the Jack-in-the-box with a stride of $9$ and $20$ patterns respectively. We captured the Jack-in-the-box by placing it horizontally to maximise its occupied field-of-view. Both videos are also included in the \href{https://wisionlab.cs.wisc.edu/project/single-photon-sl/}{teaser video} which provides an overview of our paper.

\input{supplementary_sections/figures/dynamic_depth_maps}

%% file: supplementary_sections/figures/dark_objects.tex
\begin{figure}[!h]
    \centering
    \includegraphics[width=0.5\textwidth]{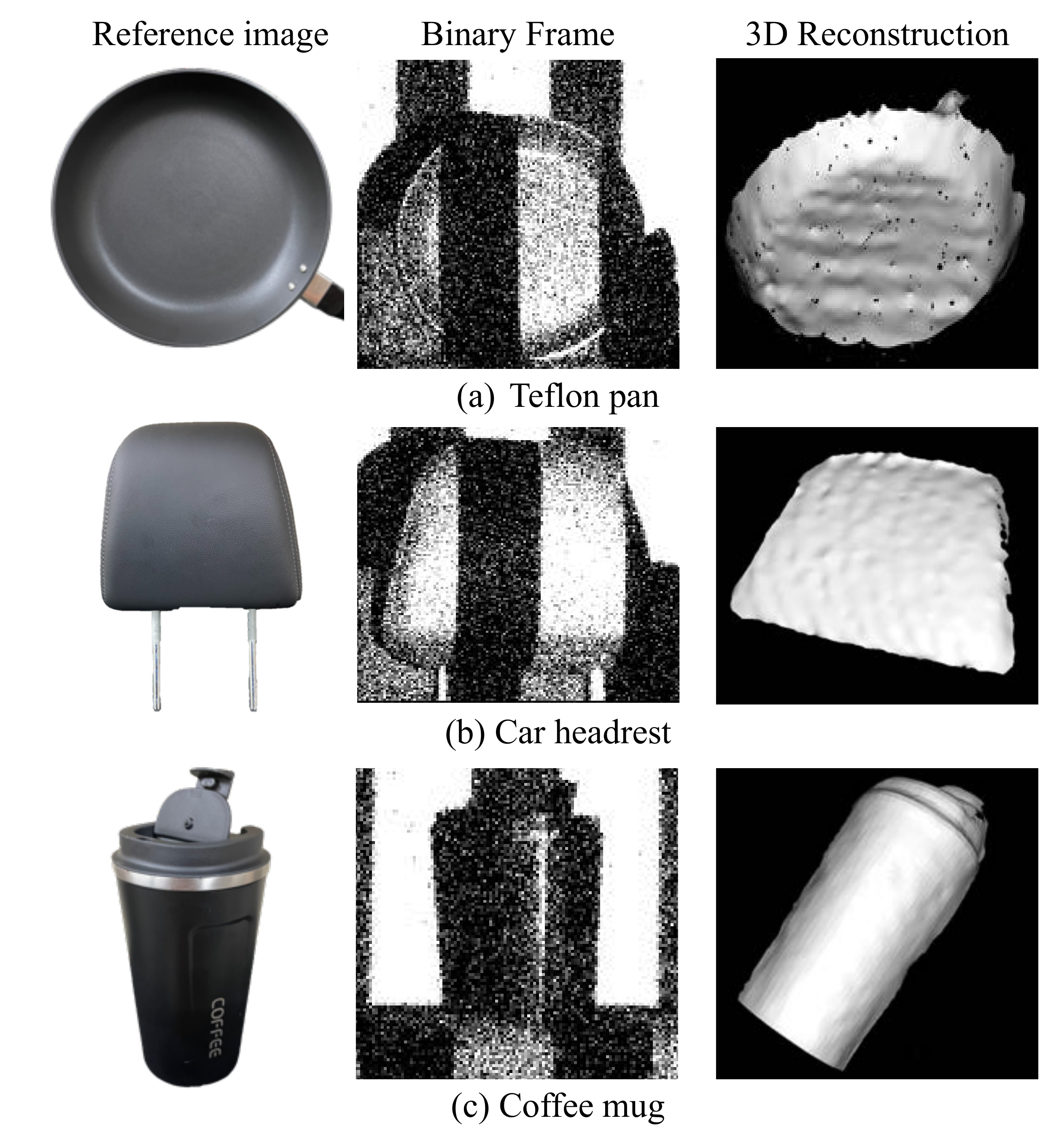}
    \caption{\textbf{More 3D reconstructions of low-albedo scenes,} using Hybrid $(n=255)$ at $40$ FPS.}
    \label{fig:supp-dark}
\end{figure}

%% file: supplementary_sections/figures/dynamic_depth_maps.tex
\begin{figure}[htp]
    \centering
    \includegraphics[width=0.95\textwidth]{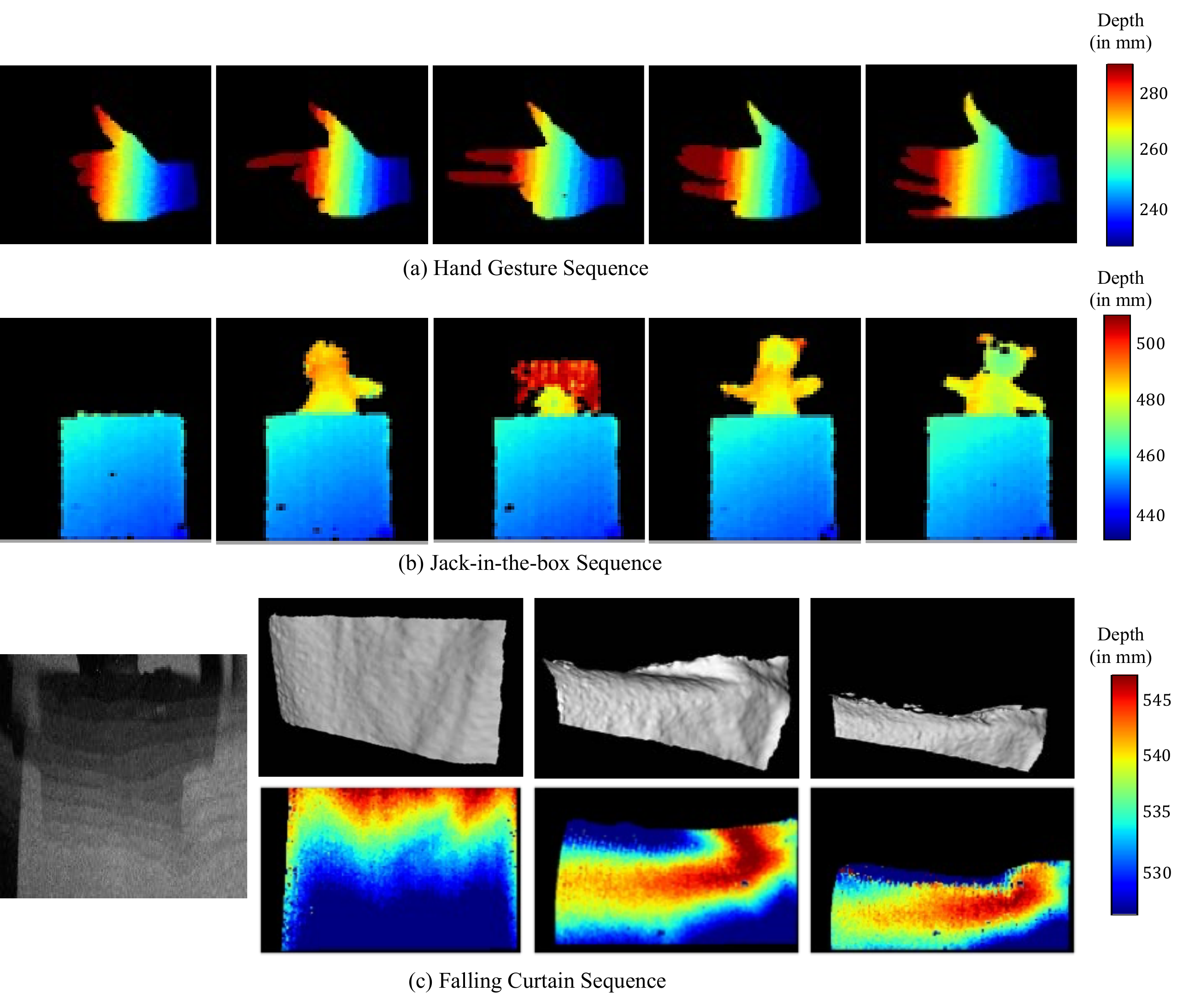}
    \caption{\textbf{Additional dynamic scenes imaged by Single-Photon SL.} The hand gesture sequence was captured using Hybrid $(n=255)$ at $75$ FPS, while the Jack-in-the-box and falling curtain sequences were both acquired using Hybrid $(n=31)$ at $450$ FPS.}
    \label{fig:supp-dyn}
\end{figure}